\pdfoutput=1
\documentclass{article}
\usepackage{ijcai17}
\usepackage{times}
\usepackage[english]{babel}
\usepackage{mathtools}
\usepackage{amssymb}
\usepackage{amsthm}
\usepackage{enumitem}
\usepackage{microtype}
\usepackage{multirow}
\usepackage{thmtools}
\usepackage{textcomp}
\usepackage{appendix}
\usepackage{titletoc}

\title{A Reasoning System for a First-Order Logic of Limited Belief\thanks{This is an extended version of the IJCAI 2017 paper [Schwering, 2017] with proofs in the appendix.}}
\author{Christoph Schwering \\ School of Computer Science and Engineering \\ The University of New South Wales \\ Sydney NSW 2052, Australia \\ {c.schwering@unsw.edu.au}}

\renewcommand\L{\ensuremath{\mathcal{L}}}
\newcommand\LL{\ensuremath{\mathcal{L\mspace{-1mu}L}}}
\newcommand{\plus}{\textsuperscript{+}}
\newcommand\Limbo{\textsc{Limbo}}
\newcommand\limbo{\textsc{limbo}}
\newcommand\e{\mathbin{=}}
\newcommand\n{\mathbin{\neq}}
\newcommand\fa[1]{\forall #1 \mspace{1mu}}
\newcommand\ex[1]{\exists #1 \mspace{1mu}}
\newcommand\limp{\supset}
\newcommand\entails{\models}
\newcommand\modelss{\mathrel{{\mathrel|\joinrel\approx}}}
\newcommand\entailss{\modelss}
\newcommand\Names{\mathcal{N}}
\newcommand\NamesP[1]{\mathcal{N}^{\text+#1}}
\newcommand\NamesPV[1]{\mathcal{N}^{\text+\mathsf{v}\text+#1}}
\newcommand\Terms{\mathcal{T}}
\newcommand\Vars{\mathcal{V}}
\newcommand\G{\mathbf{G}\mspace{1mu}}
\newcommand\K[1][]{\mathbf{K}_{#1}\mspace{1mu}}
\newcommand\M[1][]{\mathbf{M}_{#1}\mspace{1mu}}
\newcommand\OO{\mathbf{O}\mspace{1mu}}
\newcommand\Lmod[1][]{\mathbf{L}_{#1}\mspace{1mu}}
\newcommand\Min[1]{#1\textsuperscript{\textminus}}
\newcommand\Max[1]{#1\textsuperscript{+}}
\newcommand\UP[2][]{\mathsf{UP}^{#1}\ifx\uchyph#2\uchyph\else(#2)\fi}
\newcommand\UPS[2][]{\mathsf{UPS}^{#1}\ifx\uchyph#2\uchyph\else(#2)\fi}
\newcommand\VP[2][]{\Max{\mathsf{UP}^{#1}}\ifx\uchyph#2\uchyph\else(#2)\fi}
\newcommand\WP[2][]{\Min{\mathsf{UP}^{#1}}\ifx\uchyph#2\uchyph\else(#2)\fi}
\newcommand\gnd{\mathsf{gnd}}

\newcommand\BigO{\mathcal{O}}
\newcommand\RES[2]{\mathsf{RES}\ifx\uchyph#1#2\uchyph\else[#1,#2]\fi}
\newcommand\RED[2]{\|#2\|_{\ifx\uchyph#1\uchyph\else#1\fi}}
\newcommand\TRUE{{\normalfont\textsc{true}}}

\newcommand\stdname[1]{{\textrm{#1}}}
\newcommand\func[1]{{\textrm{#1}}}
\newcommand\Sally{\stdname{Sally}}
\newcommand\Frank{\stdname{Frank}}
\newcommand\Fred{\stdname{Fred}}
\newcommand\T{\top}
\newcommand\fatherOf{\func{fatherOf}}
\newcommand\rich{\func{rich}}
\newcommand\Rich{\func{Rich}}
\newcommand{\sq}[3]{\mbox{$\medmuskip=#1mu\thickmuskip=#2mu\displaystyle#3$}} 
\makeatletter
\newcommand\thmlinebreak{\leavevmode\@beginparpenalty=10000}
\makeatother
\newenvironment{itemizethm}[1][]{%
    \thmlinebreak%
    \begin{itemize}[#1]
}{%
    \end{itemize}%
}%
\newenvironment{enumeratethm}{%
    \thmlinebreak%
    \begin{enumerate}[label=(\roman*),ref=(\roman*),leftmargin=*]
}{%
    \end{enumerate}%
}%
\declaretheoremstyle[headpunct=\ ,bodyfont=\itshape]{thm}
\declaretheoremstyle[headpunct=\ ,bodyfont=\normalfont]{defi}
\declaretheoremstyle[headpunct=.\ ,headfont=\normalfont\itshape,qed=\qedsymbol]{pf}
\declaretheorem[style=thm,name=Theorem]{thm}
\declaretheorem[style=thm,name=Proposition,sibling=thm]{prop}
\declaretheorem[style=thm,name=Lemma,sibling=thm]{lem}
\declaretheorem[style=thm,name=Corollary,sibling=thm]{cor}
\declaretheorem[style=defi,name=Definition,sibling=thm]{defi}
\declaretheorem[style=defi,name=Remark,sibling=thm]{rem}
\declaretheorem[name={Proof},style=pf,numbered=no]{pf}
\newenvironment{repthm}[2][]{\let\thethmbak\thethm\def\thethm{\ref{#2}}\begin{thm}[{#1}]}{\end{thm}\let\thethm\thethmbak\addtocounter{thm}{-1}}
\newenvironment{repcor}[2][]{\let\thecorbak\thecor\def\thecor{\ref{#2}}\begin{cor}[{#1}]}{\end{cor}\let\thecor\thecorbak\addtocounter{cor}{-1}}

\SetEnumitemKey{sub}{leftmargin=1em,nosep}
\SetEnumitemKey{nolabel}{label={}}
\begin{document}
\thispagestyle{plain}
\pagestyle{plain}

\maketitle
\nocite{Schwering:LimboIJCAI}

\begin{abstract}
Logics of limited belief aim at enabling computationally feasible reasoning in highly expressive representation languages.
These languages are often dialects of first-order logic with a weaker form of logical entailment that keeps reasoning decidable or even tractable.
While a number of such logics have been proposed in the past, they tend to remain for theoretical analysis only and their practical relevance is very limited.
In this paper, we aim to go beyond the theory.
Building on earlier work by Liu, Lakemeyer, and Levesque, we develop a logic of limited belief that is highly expressive while remaining decidable in the first-order and tractable in the propositional case and exhibits some characteristics that make it attractive for an implementation.
We introduce a reasoning system that employs this logic as representation language and present experimental results that showcase the benefit of limited belief.
\end{abstract}

\section{Introduction} \label{sec:intro}

Dealing with incomplete knowledge is one of the longstanding aims of research in Knowledge Representation and Reasoning. 
Incompleteness often demands highly expressive languages to be represented accurately.
For instance, the statement \begin{center}``I don't know who Sally's father is, but I know he's rich''\end{center} involves an individual (Sally's father) and both knowns (he's rich) and unknowns (his identity) about him.
From the representational point of view, first-order and modal logics are excellent tools to formalise such statements.
However, reasoning in classical first-order logic very quickly gets undecidable: an existential quantifier and two unary functions with equality can be enough to make validity an undecidable problem \cite{Boerger:DecisionProblem}.

One way to get around undecidability of first-order reasoning is through models of \emph{limited belief}.\footnote{We use the terms knowledge and belief interchangeably.}
Inspired by natural agents, the idea behind limited belief is to give up the property of \emph{logical omniscience} \cite{Hintikka:Omniscience}.
This separates limited belief from other approaches to decidable reasoning like the classical prefix-vocabulary classes \cite{Boerger:DecisionProblem} or description logics \cite{Baader:DLHandbook}, but relates to approaches of approximate reasoning like \cite{Agostino:InformationalView}.
While a number of models of limited belief have been proposed in the past \cite{Konolige:DeductiveBelief,Vardi:Omniscience,Fagin:Limited,Levesque:Implicit,PatelSchneider:Implicit,Lakemeyer:Implicit,Delgrande:Implicit}, these approaches can be criticised for either being too fine-grained or overly weakening the entailment relation and thus even ruling out the most basic cases of modus ponens.

A more recent proposal for limited belief is due to Liu, Lakemeyer, and Levesque \shortcite{LLL:SL}.
Their logic is equipped with a perspicuous semantics based on subsumption, unit propagation, and case splitting,\footnote{Unit propagation of a clause $c$ with a literal $\ell$ means to remove all literals from $c$ that are complementary to $\ell$. Subsumption refers to inferring a clause $c_2$ from a clause $c_1$ when every literal in $c_1$ subsumes a literal in $c_2$. A case split means to branch on all possible values a literal or term can take.} and keeps the computational complexity under control by stratifying beliefs in \emph{levels}:
level 0 comprises only the explicit beliefs; every following level draws additional inferences by doing another case split.
Every query specifies at which belief level it shall be evaluated, and thus controls how much effort should be spent on proving it.
The rationale behind this technique of limiting belief by case splits is the hypothesis that in many practical reasoning tasks few case splits -- perhaps no more than one or two -- suffice.

Let us consider a brief example to illustrate the idea.
Suppose we have the following knowledge base (KB):
\begin{align*}
& (\fatherOf(\Sally) \e \Frank \lor \fatherOf(\Sally) \e \Fred) \land{}\\
& \fa x (\fatherOf(\Sally) \n x \lor \Rich(x)) \text.
\end{align*}
At level 0 the agent believes these clauses, but does not draw any meaningful inferences from them yet.
For instance, $\Rich(\Frank) \lor \Rich(\Fred)$, while entailed in classical logic, is not believed at level 0.
It can however be inferred by splitting the cases for Sally's potential fathers:
\begin{itemize}
\item if $\fatherOf(\Sally) \e \Frank$, then we obtain $\Rich(\Frank)$ by unit propagation with the second clause from the KB;
\item if $\fatherOf(\Sally) \e \Fred$, then analogously $\Rich(\Fred)$;
\item if $\fatherOf(\Sally)$ is anyone else, then unit propagation with the first clause in the KB yields the empty clause.
\end{itemize}
Either of the three cases subsumes $\Rich(\Frank) \lor \Rich(\Fred)$.
Hence, $\Rich(\Frank) \lor \Rich(\Fred)$ is believed at level 1.

Several variants of Liu, Lakemeyer, and Levesque's original theory have been developed \cite{LL:LB,LL:ESL,LL:LBF,Klassen:Limited,SL:BOL}; they include concepts like introspection, actions, functions, or conditional beliefs.
Despite this progress, the framework has remained a purely theoretical one without practical applications.

In this paper, we want to bring their approach to limited belief to practice.
We make two contributions to this end:
\begin{itemize}
\item Firstly, we devise a logic of limited belief that unifies some concepts from the earlier proposals and adds several features to make it more attractive for practical use.
    The result is a sorted first-order logic with functions and equality and two introspective belief operators, one for knowledge and one for what is considered possible. 
    While closely related to the earlier proposals, many technical details in this new language have been changed with practical considerations in mind.

\item Secondly, we present a reasoning system that uses this logic as representation language.
    Our evaluation is twofold.
    Besides modelling toy domains to showcase the language's expressivity, we have also tested the system's performance with two popular puzzle games, Sudoku and Minesweeper.
    The results affirm the hypothesis that small belief levels often suffice to achieve good results.
\end{itemize}


The paper is organised as follows.
In Section~\ref{sec:unlimited} we introduce an (omniscient) logic of knowledge.
Based on this logic, we then develop a logic of limited belief in Section~\ref{sec:limited}.
In Section~\ref{sec:system} we sketch a decision procedure for limited belief, discuss an implementation of this system, and present experiment results.

\section{A Logic of Knowledge} \label{sec:unlimited}

The logic we present in this section is a variant of Levesque's logic of only-knowing \cite{Levesque:AllIKnow} and will serve as a reference for the logic of limited belief in the next section.
We refer to the logic from this section as \L.

\subsection{The Language} \label{sec:l:lang}

The language of \L\ is a sorted first-order dialect with functions, equality, standard names, and three epistemic modalities.
With minor modifications, this language will also be the language used for the logic of limited belief \LL.

We assume an infinite supply of \emph{sorts}, and for each sort we assume an infinite number of \emph{variables}, \emph{function symbols} of every arity $j \geq 0$, and \emph{standard names} (or \emph{names} for short).
Standard names serve as special constants that satisfy the unique name assumption and an infinitary version of domain closure.
The set of \emph{terms} (of a sort $s$) contains all variables (of sort $s$) and names (of sort $s$) as well as all functions $f(t_1,\ldots,t_j)$ where $f$ is a $j$-ary function symbol (of sort $s$) and every $t_i$ is a variable or a name (not necessarily of sort $s$).
A \emph{literal} is an expression of the form $t_1 \e t_2$ or $\neg t_1 \e t_2$ where $t_1$ is a variable, name, or function, and $t_2$ is a variable or a name.
The set of \emph{formulas} is the least set that contains all literals and all expressions $\neg \alpha$, $(\alpha \lor \beta)$, $\ex x \alpha$, $\K \alpha$, $\M \alpha$, and $\OO \alpha$ where $\alpha$ and $\beta$ are formulas and $x$ is a variable.

Intuitively, $\K \alpha$ reads as ``$\alpha$ is known,'' $\M \alpha$ as ``$\alpha$ is considered possible,'' and $\OO \alpha$ as ``$\alpha$ is \emph{all} that is known.''
We say a formula is \emph{objective} when it mentions no belief operator, and \emph{subjective} when it mentions no function outside of a belief operator.
Only-knowing is particularly useful to capture the meaning of a knowledge base.
As we only consider objective knowledge bases in this paper, we restrict ourselves from now on to objective $\phi$ in $\OO \phi$.

For convenience, we use the usual abbreviations $\n$, $\land$, $\forall$, $\limp$, and $\equiv$, and we sometimes omit brackets to ease readability.

Some differences between our language and traditional first-order languages are apparent: our language features no predicates; functions cannot be nested; only the left-hand side of a literal may be a function.
These restrictions will prove helpful for the implementation of a reasoning system. 
We remark that none of these restrictions means a limitation of expressivity: a predicate $P(t_1,\ldots,t_j)$ can be simulated by a literal $p(t_1,\ldots,t_j) \e \T$ where $\T$ is some standard name chosen to represent truth, and nested functions and literals with a function on the right-hand side can be flattened by introducing a new variable -- these transformations preserve equivalence.

As an example, we formalise the introductory statement about Sally's father, who is rich but unknown to the agent:
\begin{align*}
\K \ex x \big( & \fatherOf(\Sally) \e x \land \rich(x) \e \T \land{}\\
& \M \fatherOf(\Sally) \n x \big) \text,
\end{align*}
where $\Sally$ is a standard name of the same sort (say, `human') as $x$ and $\fatherOf$, and $\T$ is a name of the same sort (say, `Boolean') as $\rich$.
Quantifying $x$ into the modal context $\M$ expresses that the father's identity is unknown.
The next subsection gives a semantic justification to this interpretation.

\subsection{The Semantics} \label{sec:l:sem}

The semantics is based on possible worlds.
We call a term $f(t_1,\ldots,t_j)$ \emph{primitive} when all $t_i$ are standard names.
Let $\Names$ and $\Terms$ be the sets of all names and primitive terms, respectively.
To denote the names or primitive terms that occur in a formula $\alpha$, we write $\Names(\alpha)$ and $\Terms(\alpha)$, respectively, and analogously for sets of formulas.
To include only terms of the same sort as $t$, we write $\Names_t$ and $\Terms_t$.
A \emph{world} $w : \Terms \rightarrow \Names$ is a sort-preserving mapping from primitive terms to standard names, that is, $w(t) \in \Names_t$ for every primitive term $t$.
A \emph{sentence} is a formula without free variables.
We denote by $\alpha^x_t$ the result of substituting $t$ for all free occurrences of the variable $x$ in $\alpha$.

Truth of a sentence $\alpha$ is defined w.r.t.\ a world $w$ and a set of possible worlds $e$ as follows:
\begin{enumerate}
\item \label{sem:l:eq} $e,w \models t \e n$ iff
    \begin{itemize}[sub]
    \item $t$ and $n$ are identical names\phantom{$w()$} \; if $t$ is a name;
    \item $w(t)$ and $n$ are identical names \; otherwise;
    \end{itemize}
\item \label{sem:l:neg} $e,w \models \neg \alpha$ iff $e,w \not\models \alpha$;
\item \label{sem:l:or} $e,w \models (\alpha \lor \beta)$ iff $e,w \models \alpha$ or $e,w \models \beta$;
\item \label{sem:l:ex} $e,w \models \ex x \alpha$ iff
    $e,w \models \alpha^x_n$ for some $n \in \Names_x$;
\item \label{sem:l:k} $e,w \models \K \alpha$ iff $e,w' \models \alpha$ for all $w' \in e$;
\item \label{sem:l:m} $e,w \models \M \alpha$ iff $e,w' \models \alpha$ for some $w' \in e$;
\item \label{sem:l:o} $e,w \models \OO \phi$ iff $e = \{w' \mid e,w' \models \phi\}$.
\end{enumerate}

Unlike classical first-order logic, this semantics handles quantification by substitution of names.
Standard names thus effectively serve as a fixed, countably infinite universe of discourse.
See \cite{Levesque:Foundations} for a discussion why this is no effective limitation for our purposes.

$\OO \phi$ has the effect of $\K \phi$ and additionally requires the set of possible worlds $e$ to be maximal.
Hence the agent knows $\phi$ but considers possible everything else provided it is consistent with $\phi$.
In other words, everything that is not a consequence of $\phi$ is unknown, for its negation is consistent with $\phi$. 
That way, $\OO \phi$ captures that $\phi$ and only $\phi$ is known.

As usual, a sentence $\alpha$ \emph{entails} another sentence $\beta$, written $\alpha \entails \beta$, when $e,w \models \alpha$ implies $e,w \models \beta$ for all $e,w$.
A sentence $\alpha$ is \emph{valid}, written $\entails \alpha$, when $e,w \models \alpha$ for all $e,w$.

We omit a deeper analysis except to note that $\K$ is a K45 operator \cite{Fagin:Knowledge} and the following equivalences:
\begin{prop} \label{prop:l}
\begin{enumeratethm}
\item $\entails \K \alpha \equiv \neg \M \neg \alpha$;
\item $\entails \fa x \K \alpha \equiv \K \fa x \alpha$ and
      $\entails \K \alpha \land \K \beta \equiv \K (\alpha \land \beta)$;
\item $\entails \ex x \M \alpha \equiv \M \ex x \alpha$ and
      $\entails \M \alpha \lor \M \beta \equiv \M (\alpha \lor \beta)$;
\item $\entails \OO \phi \limp \K \phi$.
\end{enumeratethm}
\end{prop}

To familiarise ourselves with the logic, we show that the query formalised at the end of Section~\ref{sec:l:lang} is entailed by
\begin{align*}
\OO \big(
    & (\fatherOf(\Sally) \e \Frank \lor \fatherOf(\Sally) \e \Fred) \land{}\\
    & \fa x (\fatherOf(\Sally) \n x \lor \rich(x) \e \T) \big)
\end{align*}
where $\Frank$ and $\Fred$ are names of sort `human.'
Let $\phi$ denote the sentence within $\OO$.
By Rule~\ref{sem:l:o}, $e = \{w \mid e,w \models \phi\}$ is the only set of worlds that satisfies $\OO \phi$, so proving the entailment reduces to model checking for $e$.
By assumption, for every $w \in e$, $w(\fatherOf(\Sally)) \in \{\Frank,\Fred\}$.
Suppose $w(\fatherOf(\Sally)) = \Frank$; the case for $\Fred$ is analogous.
By assumption, $w(\rich(\Frank)) = \T$, so it only remains to be shown that $e,w \models \M \fatherOf(\Sally) \n \Frank$, which holds because there are $w' \in e$ with $w'(\fatherOf(\Sally)) = \Fred$.

\section{A Logic of Limited Belief} \label{sec:limited}

We now introduce the logic \LL, the limited counterpart of \L.

\subsection{The Language} \label{sec:ll:lang}

The language of \LL\ follows the rules from \L\ with the following modifications: the expressions $\K \alpha$ and $\M \alpha$ are replaced with $\K[k] \alpha$ and $\M[k] \alpha$ where $k \geq 0$ is a natural number, and the expression $\G \alpha$ is added to the language.
We read $\K[k] \alpha$ and $\M[k] \alpha$ as ``$\alpha$ is believed at level $k$'' and ``$\alpha$ is considered possible at level $k$,'' respectively, and the new expression $\G \alpha$ intuitively means ``assuming the knowledge base is consistent, $\alpha$ is true.''
Guaranteeing consistency is motivated by practical applications where it often may reduce the computational cost of reasoning.

\subsection{The Semantics} \label{sec:ll:sem}

In \LL, sets of clauses take over from sets of possible worlds as the semantic primitive that models belief.
Intuitively, these clauses will represent the agent's explicit knowledge, like $\fatherOf(\Sally) \e \Frank \lor \fatherOf(\Sally) \e \Fred$ and $\fa{x} (\fatherOf(\Sally) \n x \lor \rich(x) \e \T)$ in our running example.
By means of case splitting and unit propagation then further inferences can be drawn from these clauses.
Before we can formalise this, we need to introduce some terminology.

We call a literal \emph{ground} when it contains no variables.
Recall that a primitive term is one of the form $f(n_1,\ldots,n_j)$ where the $n_i$ are names.
Therefore every ground literal is of form $n \e n'$ or $n \n n'$ or $f(n_1,\ldots,n_j) \e n$ or $f(n_1,\ldots,n_j) \n n$ for names $n_i, n, n'$. 

A literal is \emph{valid} when it is of the form $t \e t$, or $n \n n'$ for distinct names $n, n'$, or $t \n t'$ for terms $t, t'$ of distinct sorts.
%
A literal $\ell_1$ \emph{subsumes} a literal $\ell_2$ when $\ell_1,\ell_2$ are identical or $\ell_1,\ell_2$ are of the form $t \e n_1$ and $t \n n_2$ for distinct names $n_1, n_2$.
Two literals $\ell_1, \ell_2$ are \emph{complementary} when $\ell_1,\ell_2$ are of the form $t \e t'$ and $t \n t'$ (or vice versa), or $\ell_1,\ell_2$ are of the form $t \e n_1$ and $t \e n_2$ for distinct names $n_1, n_2$.

A \emph{clause} is a finite set of literals.
A clause with a single literal is a \emph{unit clause}.
We abuse notation and identify non-empty clauses $\{\ell_1,\ldots,\ell_j\}$ with formulas $(\ell_1 \lor \ldots \lor \ell_j)$.
The above terminology for literals carries over to clauses as follows.
A clause is \emph{valid} when it contains a valid literal, or a literal $t \e t'$ and its negation $t \n t'$, or two literals $t \n n_1$ and $t \n n_2$ for distinct names $n_1, n_2$.
A clause $c_1$ \emph{subsumes} a clause $c_2$ if every literal $\ell_1 \in c_1$ subsumes a literal $\ell_2 \in c_2$.
The \emph{unit propagation} of a clause $c$ with a literal $\ell$ is the clause obtained by removing from $c$ all literals that are complementary to $\ell$.

A \emph{setup} is a set of ground clauses.
We write $\UP{s}$ to denote the closure of $s$ with all valid literals under unit propagation:
\begin{itemize}
\item if $c \in s$, then $c \in \UP{s}$;
\item if $\ell$ is a valid literal, then $\ell \in \UP{s}$;
\item if $c, \ell \in \UP{s}$ and $c'$ is the unit propagation of $c$ with $\ell$, then $c' \in \UP{s}$.
\end{itemize}
We write $\VP{s}$ to denote the result of adding to $\UP{s}$ all valid clauses and all clauses that are subsumed by some clause in $\UP{s}$.
Similarly, $\WP{s}$ shall denote the setup obtained by removing from $\UP{s}$ all valid clauses and all clauses subsumed by some other clause in $\VP{s}$.

Truth of a sentence $\alpha$ in \LL, written $s_0,s,v \modelss \alpha$, is defined w.r.t.\ two setups $s_0, s$ and a set of unit clauses $v$.
The purpose of having these three parameters is to deal with nested beliefs.
For the objective part of the semantics, only $s$ is relevant:
\begin{enumerate}[series=semll]
\item \label{sem:ll:lit} $s_0,s,v \modelss \ell$ iff
    $\ell \in \VP{s}$ \; if $\ell$ is a literal;
\item \label{sem:ll:or} $s_0,s,v \modelss (\alpha \lor \beta)$ iff
    \begin{itemize}[sub]
    \item $(\alpha \lor \beta) \in \VP{s}$ \qquad\quad\;\;\ \, if $(\alpha \lor \beta)$ is a clause;
    \item $s_0,s,v \modelss \alpha$ or $s_0,s,v \modelss \beta$ \; otherwise;
    \end{itemize}
\item \label{sem:ll:and} $s_0,s,v \modelss \neg (\alpha \lor \beta)$ iff $s_0,s,v \modelss \neg \alpha$ and $s_0,s,v \modelss \neg \beta$;
\item \label{sem:ll:ex} $s_0,s,v \modelss \ex x \alpha$ iff $s_0,s,v \modelss \alpha^x_n$ for some $n \in \Names_x$;
\item \label{sem:ll:fa} $s_0,s,v \modelss \neg \ex x \alpha$ iff $s_0,s,v \modelss \neg \alpha^x_n$ for every $n \in \Names_x$;
\item \label{sem:ll:negneg} $s_0,s,v \modelss \neg \neg \alpha$ iff $s_0,s,v \modelss \alpha$.
\end{enumerate}
Note how negation is handled by rules for $(t_1 \n t_2)$, $\neg (\alpha \lor \beta)$, $\neg \ex x \alpha$, $\neg \neg \alpha$.
A rule $s_0,s,v \modelss \neg \alpha$ iff $s_0,s,v \not\modelss \alpha$ would be unsound, as \LL\ is incomplete w.r.t.\ \L\ (as we shall see).

We proceed with the semantics of $\K[k] \alpha$.
The idea is that $k$ case splits can be made first, before $\alpha$ is evaluated.
A case split means to select some term (say, $\fatherOf(\Sally)$) and branch (conjunctively) on the values it could take (namely all standard names of the right sort, such as $\Frank$ and $\Fred$).
To preserve soundness of introspection, the effect of case splits must not spread into nested beliefs.
This is why we need to carefully manage three parameters $s_0,s,v$.
Intuitively, $s_0$ is the ``original'' setup without split literals, and $v$ ``stores'' the split literals.
Once the number of case splits is exhausted, $s_0 \cup v$ takes the place of $s$, so that the objective subformulas of $\alpha$ are interpreted by $s_0 \cup v$, whereas the subjective subformulas of $\alpha$ are interpreted by $s_0$ (plus future splits from the nested belief operators).
We say a setup $s$ is \emph{obviously inconsistent} when $\UP{s}$ contains the empty clause.
In this special case, which corresponds to the empty set of worlds in the possible-worlds semantics, everything is known.
The semantics of knowledge formalises this idea as follows:
\begin{enumerate}[resume*=semll]
\item \label{sem:ll:k0} $s_0,s,v \modelss \K[0] \alpha$ iff
    \begin{itemize}[sub]
    \item $s_0 \cup v$ is obviously inconsistent, or
    \item $s_0,s_0 \cup v,\emptyset \modelss \alpha$;
    \end{itemize}
\item \label{sem:ll:kk} $s_0,s,v \modelss \K[k+1] \alpha$ iff
    \begin{itemize}[sub,nolabel]
    \item for some $t \in \Terms$ and every $n \in \Names_t$,
    \item $s_0,s,v \cup \{t \e n\} \modelss \K[k] \alpha$;
    \end{itemize}
\item \label{sem:ll:negk} $s_0,s,v \modelss \neg \K[k] \alpha$ iff $s_0,s,v \not\modelss \K[k] \alpha$.
\end{enumerate}

Similarly, the idea behind $\M[k] \alpha$ is to fix the value of certain terms in order to show that the setup is consistent with $\alpha$.
Intuitively this means that we want to approximate a possible world, that is, an assignment of terms to names, that satisfies $\alpha$.
Often we want to fix not just a single term, but a series of terms with a common pattern, for instance, $f(n) \e n$ for all $n$.
To this end, we say two literals $\ell_1,\ell_2$ are \emph{isomorphic} when there is a bijection $\ast: \Names \rightarrow \Names$ that swaps standard names in a sort-preserving way so that $\ell_1$ and $\ell_2^\ast$ are identical, and define $\sq{1mu plus 1mu minus 1}{2mu plus 1mu minus 1}{{v \uplus_s \ell_1} = v \cup \{\ell_2 \mid \text{$\ell_1, \ell_2$ are isomorphic and $\neg \ell_2 \notin \VP{{s \cup v}}$}\}}$.
In English: $v \uplus_s \ell_1$ adds to $v$ every literal that is isomorphic to $\ell_1$ and not obviously inconsistent with the setup $s \cup v$.
Furthermore, we need to take care that after fixing these values the setup is not potentially inconsistent.
We say a setup $s$ is \emph{potentially inconsistent} when it is obviously inconsistent or when the set $\{\ell \mid \ell \in c \in \WP{s}\}$ of all literals in $\WP{s}$ contains two complementary literals, or a literal $t \e n$ for $n \notin \Names_t$, or all literals $t \n n$ for $n \in \Names_t$ for some primitive term $t$.
Note that this consistency test is intentionally naive, for the complexity of $\M[k] \alpha$ shall be bounded by $k$ alone.
The semantics of the consistency operator is then:
\begin{enumerate}[resume*=semll]
\item \label{sem:ll:m0} $s_0,s,v \modelss \M[0] \alpha$ iff
    \begin{itemize}[sub]
    \item $s_0 \cup v$ is not potentially inconsistent, and
    \item $s_0,s_0 \cup v,\emptyset \modelss \alpha$;
    \end{itemize}
\item \label{sem:ll:mk} $s_0,s,v \modelss \M[k+1] \alpha$ iff
    \begin{itemize}[sub,nolabel]
    \item for some $t \in \Terms$ and $n \in \Names_t$,
    \item $s_0,s,v \cup \{t \e n\} \modelss \M[k] \alpha$ or
    \item $s_0,s,v \uplus_{s_0} (t \e n) \modelss \M[k] \alpha$;
    \end{itemize}
\item \label{sem:ll:negm} $s_0,s,v \modelss \neg \M[k] \alpha$ iff $s_0,s,v \not\modelss \M[k] \alpha$.
\end{enumerate}

To capture that $\OO \phi$ means that $\phi$ is all the agent knows, we need to minimise the setup (modulo unit propagation and subsumption), which corresponds to the maximisation of the set of possible worlds in \L.
We hence define:
\begin{enumerate}[resume*=semll]
\item \label{sem:ll:o} $s_0,s,v \modelss \OO \phi$ iff
    \begin{itemize}[sub]
    \item $s_0,s_0,\emptyset \modelss \phi$, and
    \item $s_0,\hat{s}_0,\emptyset \not\modelss \phi$ for every $\hat{s}_0$ with $\VP{\hat{s}_0} \subsetneq \VP{s_0}$;
    \end{itemize}
\item \label{sem:ll:nego} $s_0,s,v \modelss \neg \OO \phi$ iff $s_0,s,v \not\modelss \OO \phi$.
\end{enumerate}

Lastly, we define the semantics of the $\G \alpha$ operator, which represents a guarantee that $s$ is consistent and therefore can reduce the size of $s$ to clauses potentially relevant to $\alpha$.
We denote the \emph{grounding} of $\alpha$ by $\gnd(\alpha) = \sq35{\{\beta{}^{x_1 \ldots x_j}_{n_1 \ldots n_{\smash{j}}} \mid n_i \in \Names_{x_i}\}}$ where $\beta$ is the result of removing all quantifiers from $\alpha$, and $x_1,\ldots,x_j$ are the free variables in $\beta$.
Finally, $s|_T$ is the least set such that if $c \in \WP{s}$ and $c$ mentions a term from $T$, is empty, or shares a term with another clause in $s|_T$, then $c \in s|_T$.
Then $\G \alpha$ works as follows:
\begin{enumerate}[resume*=semll]
\item \label{sem:ll:g} $s_0,s,v \modelss \G \alpha$ iff $s_0|_{\Terms(\gnd(\alpha))},s,v \modelss \alpha$;
\item \label{sem:ll:negg} $s_0,s,v \modelss \neg \G \alpha$ iff $s_0,s,v \modelss \G \neg \alpha$.
\end{enumerate}

This completes the semantics.
We write $s_0,s \modelss \alpha$ to abbreviate $s_0,s,\emptyset \modelss \alpha$.
Note that for subjective formulas $\sigma$, $s$ is irrelevant, so we may just write $s_0 \modelss \sigma$.
Analogous to $\models$ in \L, we overload $\modelss$ for \emph{entailment} and \emph{validity}.

%
%

In this paper we are mostly concerned with reasoning tasks of the form $\OO \phi \entailss \sigma$ where $\sigma$ is a subjective query and $\phi$ is a knowledge base of a special form called \emph{proper\plus}: $\phi$ is of the form $\bigwedge_i \fa{x_1} \ldots \fa{x_j} c_i$ for clauses $c_i$.

Observe that a proper\plus\ KB directly corresponds to the setup $\gnd(\phi) = \bigcup_i \gnd(c_i)$.
Also note that while existential quantifiers are disallowed in proper\plus\ KBs, they can be simulated as usual by way of Skolemisation.

Reasoning in proper\plus\ KBs is sound in \LL\ w.r.t.\ \L, provided that the query does not mention belief modalities $\K[k], \M[k]$ in a negated context.
While the first-order case is incomplete, a restricted completeness result for the propositional case will be given below.
We denote by $\sigma_\L$ the result of replacing in $\sigma$ every $\K[k], \M[k]$ with $\K, \M$.
The soundness theorem follows:

\begin{thm} \label{thm:soundness}
Let $\phi$ be proper\plus\ and $\sigma$ be subjective, without $\OO, \G$, and without negated $\K[k], \M[k]$. 
\\
Then $\OO \phi \entailss \sigma$ implies $\OO \phi \entails \sigma_\L$.
\end{thm}

Negated beliefs break soundness because of their incompleteness.
For example, $\K[k] (t \e n \lor \neg \neg t \n n)$ in general only holds for $k \geq 1$.
Hence $\neg \K[0] (t \e n \lor \neg \neg t \n n)$ comes out true, which is unsound w.r.t.\ \L.
As a consequence of this incompleteness, $\M[k] \alpha \equiv \neg \K[k] \neg \alpha$ is not a theorem in \LL.

For \emph{propositional} formulas, that is, formulas without quantifiers, high-enough belief levels are complete:

\begin{thm} \label{thm:eventual-completeness}
Let $\phi, \sigma$ be propositional, $\phi$ be proper\plus, $\sigma$ be subjective and without $\OO, \G$.
Let $\sigma_k$ be like $\sigma$ with every $\K[l], \M[l]$ replaced with $\K[k], \M[k]$.
\\
Then $\OO \phi \entails \sigma_\L$ implies that there is a $k$ such that $\OO \phi \entailss \sigma_k$.
\end{thm}

To conclude this section, let us revisit our running example.
The KB is the same as in Section~\ref{sec:l:sem}, and the modalities in the query are now indexed with belief levels:
\begin{align*}
\K[1] \ex x \big( & \fatherOf(\Sally) \e x \land \rich(x) \e \T \land{}\\
& \M[1] \fatherOf(\Sally) \n x \big) \text.
\end{align*}
By Rule~\ref{sem:ll:o}, $s_0 = \{\fatherOf(\Sally) \e \Frank \lor \fatherOf(\Sally)$ $\mathop{\e} \Fred, \fatherOf(\Sally) \n n \lor \rich(n) \e \T \mid n \in \Names_x\}$ is the unique (modulo $\VP{}$) setup that satisfies the KB.
To prove the query, by applying Rule~\ref{sem:ll:kk} we can split the term $\fatherOf(\Sally)$.
Consider $s_0 \cup \{\fatherOf(\Sally) \e \Frank\}$.
By unit propagation we obtain ${\rich(\Frank) \e \T}$, so we can choose $\Frank$ for $x$ in Rule~\ref{sem:ll:ex}, and all that remains to be shown is that $s_0 \modelss \M[1] {\fatherOf(\Sally) \n \Frank}$.
This is done by assigning $\fatherOf(\Sally) \e \Fred$ in Rule~\ref{sem:ll:mk}, and as the resulting setup $s_0 \cup \{\fatherOf(\Sally) \e \Fred\}$ is not potentially inconsistent and subsumes $\fatherOf(\Sally) \n \Frank$, the query holds.
Returning to the splitting in Rule~\ref{sem:ll:kk}, the case $s_0 \cup \{\fatherOf(\Sally) \e \Fred\}$ is analogous, and for all other $n$, the setups $s_0 \cup \{\fatherOf(\Sally) \e n\}$ are obviously inconsistent and therefore satisfy the query by Rule~\ref{sem:ll:k0}.

\section{A Reasoning System} \label{sec:system}

We now proceed to describe a decision procedure for reasoning in proper\plus\ knowledge bases, and then discuss an implementation as well as experimental results.

\subsection{Decidability}

Reasoning in proper\plus\ knowledge bases is decidable in \LL:

\begin{thm} \label{thm:decidable}
Let $\phi$ be proper\plus, $\sigma$ be subjective and without $\OO$.
Then $\OO \phi \entailss \sigma$ is decidable.
\end{thm}

We only sketch the idea here for space reasons; for the full proof see the appendix.
For the rest of this section, we use $\Lmod[k]$ as a placeholder for $\K[k]$ and $\M[k]$.

Let us first consider the case where $\sigma$ is of the form $\Lmod_k \psi$ for objective $\psi$; we will turn to nested modalities later.
As $\phi$ is proper\plus, $\gnd(\phi)$ gives us the unique (modulo $\VP{}$) setup that satisfies $\OO \phi$, so the reasoning task reduces to model checking of $\gnd(\phi)$.
An important characteristic of standard names is that, intuitively, a formula cannot distinguish the names it does not mention.
As a consequence, we can limit the grounding $\gnd(\phi)$ and quantification during the model checking to a finite number of names.
For every variable $x$ in $\phi$ or $\psi$, let $p_x$ be the maximal number of variables occurring in any subformula of $\phi$ or $\psi$ of the same sort as $x$.
It is then sufficient for the grounding and for quantification of $x$ to consider only the names $\Names_x(\phi) \cup \Names_x(\psi)$ plus $p_x + 1$ additional new names.
Given the finite grounding and quantification, splitting in Rules \ref{sem:ll:kk} and \ref{sem:ll:mk} can also be confined to finite many literals.

That way, the rules of the semantics of \LL\ can be reduced to only deal with finite structures, which immediately yields a decision procedure for $\OO \phi \entailss \Lmod[k] \psi$.

In the propositional case, this procedure is tractable:

\begin{thm} \label{thm:tractable-objective}
Let $\phi, \psi$ be propositional, $\phi$ be proper\plus, $\psi$ be objective.
Then $\OO \phi \entailss \Lmod[k] \psi$ is decidable in $\BigO(2^k (|\phi| + |\psi|)^{k+2})$.
\end{thm}

Now we turn to nested beliefs, which are handled using Levesque's representation theorem \cite{Levesque:Foundations}.
When $\psi$ mentions a free variable $x$, the idea is to replace a nested belief $\Lmod[k] \psi$ with all instances $n$ for which $\Lmod[k] \psi^x_n$ holds.
Given a proper\plus\ $\phi$, a set of primitive terms $T$, and a formula $\Lmod[k] \psi$ for objective $\psi$, we define $\RES{\phi,T}{\Lmod[k] \psi}$ as
\begin{itemize}
\item if $\psi$ mentions a free variable $x$:\\
    $\bigvee_{n \in \Names_x(\phi) \cup \Names_x(\psi)} (x \e n \land \RES{\phi,T}{\Lmod[k] \psi^x_n}) \lor{}$\\
    $\big(\bigwedge_{n \in \Names_x(\phi) \cup \Names_x(\psi)} x \n n \land \RES{\phi,T}{\Lmod[k] \psi^x_{\hat{n}}}^{\hat{n}}_x\big)$\\
    where $\hat{n} \in \Names_x \setminus (\Names_x(\phi) \cup \Names_x(\psi))$ is a some new name;
\item if $\psi$ mentions no free variables:\\
    $\TRUE$ if $\gnd(\phi)|_T \modelss \Lmod[k] \psi$, and $\neg \TRUE$ otherwise,
    where $\TRUE$ stands for $\ex x x \e x$.
\end{itemize}
In our running example, $\RES{\phi,\Terms}{\M[1] \fatherOf(\Sally) \n x}$ is $(x \e \Frank \land \TRUE) \lor (x \e \Fred \land \TRUE) \lor (x \e \Sally \land \TRUE) \lor (x \n \Frank \land x \n \Fred \land x \n \Sally \land \TRUE)$, which says that everybody is potentially \emph{not} Sally's father.

The $\RES{}{}$ operator can now be applied recursively to eliminate nested beliefs from the inside to the outside.
For proper\plus\ $\phi$, a set of terms $T$, and $\alpha$ without $\OO$, we define
\begin{itemize}
\item $\RED{\phi,T}{t \e t'}$ as $t \e t'$;
\item $\RED{\phi,T}{\neg \alpha}$ as $\neg \RED{\phi,T}{\alpha}$;
\item $\RED{\phi,T}{(\alpha \lor \beta)}$ as $(\RED{\phi,T}{\alpha} \lor \RED{\phi,T}{\beta})$;
\item $\RED{\phi,T}{\ex x \alpha}$ as $\ex x \RED{\phi,T}{\alpha}$;
\item $\RED{\phi,T}{\Lmod[k] \alpha}$ as $\RES{\phi,T}{\Lmod[k] \RED{\phi,T}{\alpha}}$;
\item $\RED{\phi,T}{\G \alpha}$ as $\RED{\phi,T \cap \Terms(\gnd(\alpha))}{\alpha}$.
\end{itemize}
Note that $\RED{}{\cdot}$ works from the inside to the outside and always returns an objective formula.
In our example, $\RED{\phi,\Terms}{\K[1] \ldots}$ first determines $\RES{\phi,\Terms}{\M[1] \fatherOf(\Sally) \n x}$, and then $\RES{\phi,\Terms}{\K[1] \ex x (\fatherOf(\Sally) \e x \land \rich(x) \e \T \land \RES{\phi,\Terms}{\M[1] \fatherOf(\Sally) \n x)}}$ evaluates to $\TRUE$.

Levesque's representation theorem (transferred to \LL) states that this reduction is correct:

\begin{thm} \label{thm:representation}
Let $\phi$ be proper\plus, $\sigma$ be subjective and without $\OO$.
Then $\OO \phi \entailss \sigma$ iff $\entailss \RED{\phi,\Terms}{\sigma}$.
\end{thm}

It follows that propositional reasoning is tractable:

\begin{cor} \label{cor:tractable}
Let $\phi, \sigma$ be propositional, $\phi$ be proper\plus, $\sigma$ be subjective, and $k \geq l$ for every $\K[l], \M[l]$ in $\sigma$.
Then $\OO \phi \entailss \sigma$ is decidable in $\BigO(2^k (|\phi| + |\sigma|)^{k+3})$.
\end{cor}

\subsection{Implementation}

The reasoning system \limbo\ implements the decision procedure sketched in the previous subsection.
\Limbo\ is written in C++ and available as open source.\footnote{The system is available {www.github.com/schwering/limbo}.}

Compared to literals in propositional logic, literals in our first-order language are relatively complex objects.
As we want to adopt SAT solving technology, care was taken to keep literal objects lightweight and efficient.
To this end, a technique called interning is used for terms, so that for every term only one copy is created and stored in a pool, and the term from then on is uniquely identified by a 31\,bit number that points to its full representation in the pool.
With this lightweight representation of terms, a literal fits into a single 64\,bit number: two 31\,bit numbers for the left- and right-hand terms, and one more bit to indicate whether the literal is an equality or inequality.
Furthermore, every term index encodes whether the term is a standard name or not.
Hence, all information that is needed for the complement or subsumption tests for two literals (as defined in Section~\ref{sec:ll:sem}) is stored directly in the literal representation, so that these tests reduce to simple bitwise operations on the literals' 64\,bit representation.
Note that this lightweight representation is supported by our strict syntactic definition of a literal, which allows a function only on the left-hand side.
An experiment showed that this lightweight representation of terms and literals speeds up complement and subsumption operations by a factor of 24 compared to the naive representation.

Complement and subsumption tests for literals are mostly used in the context of setups for determining unit propagation and subsumption.
To avoid unnecessary blowup, invalid literals in clauses as well as valid clauses in setups are not represented explicitly.
To facilitate fast unit propagation and cheap backtracking during splitting, the setup data structure uses the watched-literals scheme \cite{Gomes:SAT}.
Other SAT technologies like backjumping or clause learning are not used at the current stage.
The setup data structure also provides an operation to query the value of a primitive term, which is used to optimise subformulas of the form $\K[k] t \e x$ in queries.

Other than this special case, grounding and substitution are handled naively at the moment and offer much room for improvement.
Within the $\G$ operator the system confines the setup, which often reduces the branching factor for splitting and thus improves performance.
Note that in case of an inconsistent KB, using the $\G$ operator may violate soundness, as it may discard the inconsistent clauses from the setup.

As mentioned before, predicates can be simulated using functions and a name $\T$ that represents truth of the predicate.
Typically, $\T$ is of a specific sort (say, `Boolean'), and no other names or variables of that sort occur in the KB or query.
It is noteworthy that this representation of predicates by functions is free of overhead, because besides $\T$ only one other name needs to be considered for splitting (to represent falsity).


The system also rewrites formulas, exploiting equivalences that hold in \L\ but not in \LL\ like Proposition~\ref{prop:l}~(ii--iii), and provides syntactic sugar for nested functions and the like.


\subsection{Evaluation} \label{sec:evaluation}

\begin{table}
\newcommand\mypct{\rlap{\%}\phantom{m}}
\newcommand\mysec{\rlap{\,s}\phantom{m}}
\newcommand\mymin{\rlap{\,m}\phantom{m}}
\newcommand\mynot{\rlap{}\phantom{m}}
\footnotesize
\centering
\begin{tabular}{|l|r|r|r|r|}
\hline
& \multicolumn{1}{|c|}{NYT easy}
& \multicolumn{1}{|c|}{NYT medium}
& \multicolumn{1}{|c|}{NYT hard}
& \multicolumn{1}{|c|}{Top 1465} \\ \hline
clues   & 38.0\mynot & 24.4\mynot & 24.0\mynot & 18.0\mynot \\ \hline
level 0 & 42.8\mynot & 49.5\mynot & 44.2\mynot & 45.1\mynot \\ \hline
level 1 &  0.3\mynot &  6.6\mynot & 11.2\mynot &  9.5\mynot \\ \hline
level 2 &   --\mynot &  0.5\mynot &  1.8\mynot &  4.6\mynot \\ \hline
level 3 &   --\mynot &   --\mynot &   --\mynot &  3.1\mynot \\ \hline
level 4 &   --\mynot &   --\mynot &   --\mynot &  0.5\mynot \\ \hline
level 5 &   --\mynot &   --\mynot &   --\mynot &  0.0\rlap{1}\mynot \\ \hline
time    &  0.1\mysec &  0.8\mysec &  4.1\mysec & 49.5\mymin \\ \hline
\end{tabular}
\caption{Sudoku experiments over eight puzzles of each category from The New York Times website as well as the first 125 of the ``Top 1465'' list.
The rows show how many cells on average per game were preset or solved at belief level 1, 2, 3, 4, or 5.
The last row shows the average time per puzzle.
}
\label{table:sudoku}
\end{table}

Three sample applications were developed to evaluate the reasoning system.\footnote{Browser-based versions of these examples can be accessed at {www.cse.unsw.edu.au/\texttildelow{}cschwering/limbo}.}
For one thing, a textual user interface allows for specification of reasoning problems and has been used to model several small-scale examples, including this paper's running example, to test the system's full expressivity.
In this section, however, we focus on the application of limited belief to the games of Sudoku and Minesweeper.

Sudoku is played on a 9$\times$9 grid which is additionally divided into nine 3$\times$3 blocks.
The goal is to find a valuation of the cells such that every row, column, and 3$\times$3 block contains every value $[1,9]$ exactly once.
The difficulty depends on how many and which numbers are given as clues from the start.

In Minesweeper the goal is to explore a grid by uncovering all and only those cells that contain no mine.
When such a safe cell is uncovered, the player learns how many adjacent cells are safe, but when a mined cell is uncovered, the game is lost.
The difficulty depends on the number of mines and grid size.

Both games were played by simple agents that use the reasoning system to represent and infer knowledge about the current game state.
For Sudoku, we use a function $\func{value}(x,y) \in [1,9]$ and translate the game rules to constraints such as $y_1 \e y_2 \lor \func{value}(x,y_1) \neq \func{value}(x,y_2)$.
For Minesweeper a Boolean function $\func{isMine}(x,y)$ is used, and when a cell $(x,y)$ is uncovered, clauses are added to represent the possible valuations of $\func{isMine}(x \pm 1, y \pm 1)$.
Both agents use iterative deepening to find their next move: first, they look for a cell $(x,y)$ for which $\func{value}(x,y)$ or $\func{isMine}(x,y)$ is known at belief level 0; if none exists, they repeat the same for belief level 1; and so on, until a specified maximum level is reached.
Once a known cell is found, the corresponding information is added to the knowledge base.
In the case of Minesweeper, it is sometimes necessary to guess; we then use a naive strategy that prefers cells that are not next to an uncovered field.

\begin{table}
\newcommand\mypct{\rlap{\%}\phantom{m}}
\newcommand\mysec{\rlap{\,s}\phantom{m}}
\footnotesize
\centering
\begin{tabular}{@{}|lr|r|r|r|r|}
\hline
\multicolumn{2}{|r|}{\hspace{-1pt}max level}
& \multicolumn{1}{|c|}{8$\times$8--10}
& \multicolumn{1}{|c|}{16$\times$16--40}
& \multicolumn{1}{|c|}{16$\times$30--99}
& \multicolumn{1}{|c|}{32$\times$64--320} \\ \hline
\multirow{2}{*}{0}
& rate & 62.0\mypct & 46.0\mypct &  1.4\mypct  & 28.3\mypct \\
& time & 0.01\mysec & 0.06\mysec & 0.24\mysec & 5.08\mysec\\
\hline
\multirow{2}{*}{1}
& rate & 87.3\mypct & 84.9\mypct & 37.7\mypct & 69.8\mypct \\
& time & 0.01\mysec & 0.08\mysec & 0.43\mysec & 5.46\mysec\\
\hline
\multirow{2}{*}{2}
& rate & 87.8\mypct & 85.0\mypct & 39.1\mypct & 70.0\mypct \\
& time & 0.02\mysec & 0.10\mysec & 0.64\mysec & 5.60\mysec\\
\hline
\multirow{2}{*}{3}
& rate & 87.8\mypct & 85.0\mypct & 39.1\mypct & 70.0\mypct \\
& time & 0.07\mysec & 0.25\mysec & 4.94\mysec & 5.90\mysec\\
\hline
\end{tabular}
\caption{Minesweeper experiments over 1000 randomised runs of different game configurations, where W$\times$H--M means M mines on a W$\times$H grid.
The rows contain results for different maximum belief levels used by the reasoner to figure out whether cells are safe or not.
Rate denotes the percentage of won games, time is the average time per game.}
\label{table:minesweeper}
\end{table}

While both games do not require much expressivity to be modelled, they are nevertheless interesting applications of limited belief because they are known to be computationally hard -- Sudoku on N$\times$N grids is NP-complete \cite{Takayuki:Sudoku}, Minesweeper is co-NP-complete \cite{Scott:Minesweeper} -- yet often easily solved by humans.
According to the motivating hypothesis behind limited belief, a small split level should often suffice to reach human-level performance.
Indeed we find this hypothesis confirmed for both games.
The results for Sudoku in Table~\ref{table:sudoku} show that most `easy' instances are solved just by unit propagation, and the number of necessary case splits increases for `medium' and `hard.'
Significantly more effort is needed to solve games from the ``Top 1465'' list, a repository of extremely difficult Sudokus.
For Minesweeper, Table~\ref{table:minesweeper} shows that strong results are already achieved at level 1 already, and while belief level 2 increases the chance of winning by 0.5\%, there is no improvement at level 3.

The Sudoku experiments show a strong increase in runtime for higher belief levels. 
This is caused by the cells being highly connected through constraints, so that the $\G$ operator has only little effect in terms of confining the setup and relevant split terms.
In Minesweeper, by contrast, using $\G$ improves runtime considerably.
The experiments were conducted using custom implementations of both games.
Note that while the numbers for Minesweeper are competitive with those reported by \cite{Geffner:BeliefTracking,Buffet:Minesweeper}, small differences in the rules may make them incomparable.
The tests were compiled with \texttt{clang -O3} and run on an Intel Core i7-4600U CPU at 3.3\,GHz.

\section{Conclusion} \label{sec:conclusion}

We developed a practical variant of Liu, Lakemeyer, and Levesque's \shortcite{LLL:SL} theory of limited belief and introduced and evaluated \limbo, a reasoning system that implements this logic.
The system features a sorted first-order language of introspective belief with functions and equality.
The computational complexity of reasoning in this highly expressive logic is controlled through the number of allowed case splits, which keeps reasoning decidable in general and sometimes even tractable.
The motivating hypothesis behind limited belief is that often a small number of case splits is sufficient to obtain useful reasoning results; this hypothesis was confirmed in our experimental evaluation using Sudoku and Minesweeper.

A natural next step is to incorporate (limited) theories of action, belief change, and/or multiple agents into the system.
This could open up interesting applications in epistemic planning and high-level control in cognitive robotics.
Another challenge is to improve the system's performance and expressivity using SAT and ASP solving technology such as clause learning, backjumping, efficient grounding techniques, and background theories.

%


\begin{appendices}
\onecolumn
\section*{Appendix: Proofs}
\vspace{1ex}
\startcontents
\printcontents{}{1}{\subsection*{Contents}}

\subsubsection*{}
We begin with some auxiliary definitions and basic lemmas in Appendix~\ref{app:basics} and showing a unique-model property of limited only-knowing in Appendix~\ref{app:o}.
Then we proceed to show the soundness and eventual completeness results, Theorems \ref{thm:soundness} and \ref{thm:eventual-completeness}, in Appendices \ref{app:soundness} and \ref{app:eventual-completeness}, respectively.
Next, we prove the Levesque's representation theorem for limited belief as stated in Theorem~\ref{thm:representation} in Appendix~\ref{app:representation}.
Then we turn to the decidability result, Theorem~\ref{thm:decidable}, and the complexity analysis from Theorem~\ref{thm:tractable-objective} and Corollary~\ref{cor:tractable} in Appendix~\ref{app:decidable}.

All definitions are assumed as above.
Some of them are restated, and, in some cases, refined, for convenience.

\section{Basics} \label{app:basics}


\begin{defi}[Term]
The set of \emph{terms} (of a sort $s$) contains all variables and names (of sort $s$) as well as all functions $f(t_1,\ldots,t_j)$ where $f$ is a $j$-ary function symbol (of sort $s$) and every $t_i$ is a variable or a name (not necessarily of sort $s$).
We call a term $f(t_1,\ldots,t_j)$ \emph{primitive} when all $t_i$ are standard names.
We denote by $\Names$ and $\Terms$ be the sets of all names and primitive terms, respectively.
To include only terms of the same sort as $t$, we write $\Names_t$ and $\Terms_t$.
\end{defi}

\subsection{Literals}

\begin{defi}[Literal]
A \emph{literal} is an expression of the form $t_1 \e t_2$ or $\neg t_1 \e t_2$ where $t_1$ is a variable, name, or function, and $t_2$ is a variable or a name.
A literal is \emph{ground} when it contains no variables.
Hence any ground literal is of the form $n \e n'$, $n \n n'$, $t \e n$, or $t \n n$ for names $n, n'$ and a primitive term $t$.
Literals of the former two forms, $t \e n$ and $t \n n$, are called \emph{primitive} as well.
A clause is \emph{valid} when it contains a valid literal, or a literal $t \e t'$ and its negation $t \n t'$, or two literals $t \n n_1$ and $t \n n_2$ for distinct names $n_1, n_2$.
Analogously, a literal is \emph{invalid} when it is of the form $t \n t$, or $n \e n'$ for distinct names $n, n'$, or $t \e t'$ for terms $t, t'$ of distinct sorts.
\end{defi}

\begin{lem} \label{lem:literal-valid}
Let $\ell$ be ground.
Then $\ell$ is valid iff $\entails \ell$.
\end{lem}

\begin{pf}
For the only-if direction, suppose $\ell$ is valid.
When $\ell$ is of the form $t \e t$ or $n \n n'$, the result immediately follows from the semantics.
When $\ell$ is of the form $t \n t'$ for two terms of distinct sort, then $\Names_t \ni w(t) \neq w(t') \in \Names_{t'}$ since $\Names_t \cap \Names_{t'} = \emptyset$, and hence $\entails t \n t'$.

For the converse, suppose $\ell$ is not valid.
A literal can have the following forms: $n \e n$ (*), $n \e n'$, $n \n n$, $n \n n'$ (*), $t \e t$ (*), $t \e n$, $t \n n$ for same sorts, $t \n n$ for distinct sorts, for distinct $n, n'$ and a primitive $t$, all of, unless specified otherwise, perhaps of different sorts.
The cases marked (*) need no consideration as they are valid.
Clearly, $\not\entails n \e n'$ and $\not\entails n \n n$.
For $t \e n$, there is a $w$ with $w(t) \neq n$, so $\not\entails t \e n$.
For $t \n n$ of the same sort, there is a $w$ with $w(t) = n$, so $\not\entails t \n n$.
\end{pf}

\begin{lem} \label{lem:literal-invalid}
Let $\ell$ be ground.
Then $\ell$ is invalid iff $\entails \neg \ell$.
\end{lem}

\begin{pf}
For the only-if direction, suppose $\ell$ is invalid.
When $\ell$ is of the form $t \n t$ or $n \e n'$, the result immediately follows from the semantics.
When $\ell$ is of the form $t \e t'$ for two terms of distinct sort, then $\Names_t \ni w(t) \neq w(t') \in \Names_{t'}$ since $\Names_t \cap \Names_{t'} = \emptyset$, and hence $\entails \neg t \e t'$.

For the converse, suppose $\ell$ is not invalid.
A literal can have the following forms: $n \e n$, $n \e n'$ (*), $n \n n$ (*), $n \n n'$, $t \e n$ for same sorts, $t \e n$ for distinct sorts (*), $t \n t$ (*), $t \n n$, for distinct $n, n'$ and a primitive $t$, all of, unless specified otherwise, perhaps of different sorts.
The cases marked (*) need no consideration as they are invalid.
Clearly, $\not\entails \neg n \e n$ and $\not\entails \neg n \n n'$.
For $t \e n$ of the same sorts, there is a $w$ with $w(t) = n$, so $\not\entails \neg t \e n$.
For $t \n n$, there is a $w$ with $w(t) \neq n$, so $\not\entails \neg t \n n$.
\end{pf}

\begin{lem} \label{lem:literal-names-not-valid}
Let $\ell$ be not valid and the left-hand side be a name.
Then $\entails \neg \ell$.
\end{lem}

\begin{pf}
$\ell$ is of the form $n \e n$ (*), $n \n n$, $n \e n'$, or $n \n n'$ (*) for distinct, $n, n'$.
The cases marked (*) need no consideration as they are valid.
For the remaining cases, clearly $\entails \neg n \n n$ and $\entails \neg n \e n'$.
\end{pf}

\begin{defi}[Literal subsumption, unit propagation]
A literal $\ell_1$ \emph{subsumes} a literal $\ell_2$ when $\ell_1,\ell_2$ are identical or $\ell_1,\ell_2$ are of the form $t \e n_1$ and $t \n n_2$ for two distinct names $n_1, n_2$.
Two literals $\ell_1, \ell_2$ are \emph{complementary} when $\ell_1,\ell_2$ are of the form $t \e t'$ and $t \n t'$ (or vice versa), or $\ell_1,\ell_2$ are of the form $t \e n_1$ and $t \e n_2$ for distinct names $n_1, n_2$.
\end{defi}

\begin{lem} \label{lem:literal-subsumption}
Suppose $\ell_1, \ell_2$ are ground and $\ell_1$ subsumes $\ell_2$.
Then $\entails \ell_1 \limp \ell_2$.
\end{lem}

\begin{pf}
Suppose $w \models \ell_1$.
If $\ell_1, \ell_2$ are identical, the lemma holds trivially.
If $\ell_1, \ell_2$ are of the form $t \e n_1$ and $t \n n_2$ for distinct $n_1, n_2$ and $w \models t \e n_1$, then $w(t) = n_1 \n n_2$ and thus $w \models t \n t_2$.
\end{pf}

\begin{lem} \label{lem:literal-subsumption-transitive}
Suppose $\ell_1$ subsumes $\ell_2$ and $\ell_2$ subsumes $\ell_3$.
Then $\ell_1$ subsumes $\ell_3$.
\end{lem}

\begin{pf}
First suppose $\ell_1$ and $\ell_2$ are identical.
If $\ell_2$ and $\ell_3$ are identical, then $\ell_1$ subsumes $\ell_3$ trivially.
If $\ell_2$, $\ell_3$ are of the form $t \e n_2$ and $t \n n_3$ for distinct $n_2, n_3$, then $\ell_1, \ell_3$ are of the form $t \e n_2$ and $t \n n_3$, so $\ell_1$ subsumes $\ell_3$.

Now suppose $\ell_1, \ell_2$ are of the form $t \e n_1$ and $t \n n_2$ for distinct $n_1, n_2$.
If $\ell_2$ and $\ell_3$ are identical, then $\ell_1$ subsumes $\ell_3$ trivially.
The remaining case that $\ell_2$, $\ell_3$ are of the form $t \e n'_2$ and $t \n n_3$ for distinct $n'_2, n_3$ is incompatible with the assumption that $\ell_2$ is of the form $t \n n_2$.
\end{pf}

\begin{lem} \label{lem:literal-subsumption-asymmetric}
Let $\ell_1$ subsume $\ell_2$ and $\ell_2$ subsume $\ell_1$.
Then $\ell_1, \ell_2$ are identical.
\end{lem}

\begin{pf}
By assumption, $\ell_1,\ell_2$ are identical or of the form $t_1 \e n_1$ and $t_1 \n n_2$ for distinct $n_1, n_2$.
Analogously, $\ell_2,\ell_1$ are identical or of the form $t_1 \e n_1$ and $t_1 \n n_2$ for distinct $n_1, n_2$.
Hence $\ell_1, \ell_2$ must be identical.
\end{pf}

\begin{lem} \label{lem:literal-complement}
Suppose $\ell_1, \ell_2$ are ground and complementary.
Then $\ell_1 \land \ell_2$ is unsatisfiable.
\end{lem}

\begin{pf}
First $\ell_1, \ell_2$ are of the form $t \e t'$ and $t \n t'$.
Then $t'$ is a name.
If $w(t) = t'$, then $w \not\models t \n t'$.
If $w(t) \neq t'$, then $w \not\models t \e t'$.

Now suppose $\ell_1, \ell_2$ are of the form $t \e n_1$ and $t \e n_2$ for distinct $n_1, n_2$.
If $w(t) = n_1$, then $w \not\models t \e n_2$.
If $w(t) \neq n_1$, then $w \not\models t \e n_1$.
\end{pf}

\begin{lem} \label{lem:literal-subsumption-complement}
Suppose $\ell_1$ subsumes $\ell_2$, $\ell_3$ subsumes $\ell_4$, and $\ell_2, \ell_4$ are complementary.
Then $\ell_1, \ell_3$ are complementary.
\end{lem}

\begin{pf}
By assumption, $\ell_1, \ell_2$ are identical or of the form $t \e n_1$ and $t \n n_2$ for distinct $n_1, n_2$.
Analogously, $\ell_3, \ell_4$ are identical or of the form $t \e n_1$ and $t \n n_2$ for distinct $n_1, n_2$.
Moreover, $\ell_2, \ell_4$ are of the form $t \e t'$ and $t \n t'$, or $t \n t'$ and $t \e t'$, or $t \e n_1$ and $t \e n_2$ for distinct $n_1, n_2$.

First suppose $\ell_1, \ell_2$ are identical.
If $\ell_3, \ell_4$ are identical as well, the lemma holds trivially.
Otherwise, $\ell_3, \ell_4$ are of the form $t \e n_1$ and $t \n n_2$ for distinct $n_1, n_2$.
Then $\ell_2$ is of the form $t \e n_2$.
Then $\ell_1$ is of the form $t \e n_2$, too.
Then $\ell_1, \ell_3$ are of the form $t \e n_2$ and $t \e n_1$ and hence complementary.

Now suppose $\ell_1, \ell_2$ are of the form $t \e n_1$ and $t \n n_2$ for distinct $n_1, n_2$.
Then $\ell_4$ is of the form $t \e n_2$.
Then $\ell_3$ must be identical to $\ell_4$.
Then $\ell_1, \ell_3$ are of the form $t \e n_1$ and $t \e n_2$ for distinct $n_1, n_2$, and thus complementary.
\end{pf}

\begin{rem}
If $\ell_1, \ell_2$ are complementary and $\ell_2$ subsumes $\ell_3$, then $\ell_1, \ell_3$ are not necessarily complementary: $\ell_1, \ell_2$ could be $t \n n_1$, $t \e n_1$, and $\ell_2, \ell_3$ could be $t \e n_1$, $t \n n_2$.
\end{rem}

\subsection{Clauses}

\begin{defi}[Clause]
A \emph{clause} is a set of literals.
A clause with a single literal is a \emph{unit clause}.
We abuse notation and identify the non-empty clauses $\{\ell_1,\ldots,\ell_j\}$ with formulas $(\ell_1 \lor \ldots \lor \ell_j)$.
A clause is \emph{valid} when it contains a valid literal, or a literal $t \e t'$ and its negation $t \n t'$, or two literals $t \n n_1$ and $t \n n_2$ for distinct names $n_1, n_2$.
A clause is \emph{invalid} when all its literals are invalid.
\end{defi}

\begin{lem} \label{lem:clause-valid}
A ground clause is valid iff $\entails c$.
\end{lem}

\begin{pf}
For the only-if direction, there are three cases.
If a literal in $c$ is valid, then by Lemma~\ref{lem:literal-valid} the lemma holds.
If two literals $t \e t'$ and $t \n t'$ are in $c$, then $w(t) = t'$ or $w(t) \neq t'$ for every $w$ and hence $w \models c$.
If two literals $t \n n_1$ and $t \n n_2$ for distinct $n_1, n_2$ are in $c$, then either $w(t) \neq n_1$ or $w(t) \neq n_2$ for every $w$ and hence $w \models c$.

Conversely, suppose $c$ is not valid.
Then all its clauses are not valid and there are no two clauses of the form $t \e n$ and $t \n n$ or $t \n n_1$ and $t \n n_2$ for distinct $n_1, n_2$.
By Lemma~\ref{lem:literal-names-not-valid}, all literals without a primitive term on the left-hand side are unsatisfiable, so we only need to to consider the literals of the form $t \e n$ and $t \n n$.
Let $w$ be such that for every $t \n n \in c$, $w(t) = n$.
Such $w$ is well-defined because there are no two $t \n n_1, t \n n_2 \in c$ for distinct $n_1, n_2$.
Also note that for every $t \e n \in c$ there is no $t \n n' \in c$, so we can furthermore let $w(t) = \hat{n}$, where $\hat{n} \notin \{\bar{n} \mid t \e \bar{n} \in c\}$.
By construction, $w \not\models c$.
\end{pf}

\begin{defi}[Clause subsumption, unit propagation]
A clause $c_1$ \emph{subsumes} a clause $c_2$ if every literal $\ell_1 \in c_1$ subsumes a literal $\ell_2 \in c_2$.
The unit propagation of a clause $c$ with a literal $\ell$ is the clause obtained by removing from $c$ all literals that are complementary to $\ell$.
\end{defi}

\begin{lem} \label{lem:clause-subsumption}
Suppose $c_1$ subsumes $c_2$.
Then $\entails c_1 \limp c_2$.
\end{lem}

\begin{pf}
Suppose $w \models c_1$.
Then $w \models \ell_1$ for some $\ell_1 \in c_1$.
By assumption, $\ell_1$ subsumes some $\ell_2 \in c_2$.
By Lemma~\ref{lem:literal-subsumption}, $w \models \ell_2$.
Thus $w \models c_2$.
\end{pf}

\begin{lem} \label{lem:clause-subsumption-transitive}
Suppose $c_1$ subsumes $c_2$ and $c_2$ subsumes $c_3$.
Then $c_1$ subsumes $c_3$.
\end{lem}

\begin{pf}
Let $\ell_1 \in c_2$ subsume $\ell_2$, and $\ell_2$ subsume $\ell_3 \in c_3$, which exist by assumption.
By Lemma~\ref{lem:literal-subsumption-transitive} the lemma follows.
\end{pf}

\begin{lem} \label{lem:clause-subsumption-asymmetric}
Let $c_1$ subsume $c_2$ and $c_2$ subsume $c_1$.
Then $c_1, c_2$ are identical.
\end{lem}

\begin{pf}
By assumption, every $\ell_1 \in c_1$ subsumes an $\ell_2 \in c_2$, and by Lemma~\ref{lem:literal-subsumption-asymmetric}, $\ell_1, \ell_2$ are identical, so $c_1 \subseteq c_2$.
By the analogous argument, $c_2 \subseteq c_1$.
\end{pf}

\begin{lem} \label{lem:clause-unit-propagation}
Suppose $c_2$ is the unit propagation of $c_1, \ell$.
Then $\entails c_1 \land \ell \limp c_2$.
\end{lem}

\begin{pf}
Suppose $w \models c_1 \land \ell$.
Then $w \models \ell_1$ for some $\ell_1 \in c_1$.
If $\ell_1, \ell$ were complementary, then $w \not\models \ell_1 \land \ell$ by Lemma~\ref{lem:literal-complement}, which contradicts the assumption.
Thus $\ell_1, \ell$ are not complementary.
Hence $\ell_1 \in c_2$ and thus $w \models c_2$.
\end{pf}

\begin{lem} \label{lem:clause-subsumption-unit-propagation}
Suppose $c_1$ subsumes $c_2$, $\ell_1$ subsumes $\ell_2$, and $c'_1$ and $c'_2$ are the unit propagations of $c_1$ with $\ell_1$ and $c_2$ with $\ell_2$.
Then $c'_1$ subsumes $c'_2$.
\end{lem}

\begin{pf}
By assumption, every $\ell'_1 \in c_1$ subsumes an $\ell'_2 \in c_2$.
By Lemma~\ref{lem:literal-subsumption-complement}, if $\ell'_2$ is complementary to $\ell_2$, then $\ell'_1$ is complementary to $\ell_1$.
Hence, if $\ell'_1 \in c'_1$, then $\ell'_1$ is not complementary to $\ell_1$, and so the $\ell'_2 \in c_2$ which is subsumed by $\ell'_1$ is not complementary to $\ell_2$ and hence $\ell'_2 \in c'_2$.
\end{pf}

\subsection{Setups}

\begin{defi}[Setup]
A \emph{setup} is a set of ground clauses.
We write $\UP{s}$ to denote the closure of $s$ together with all valid clauses under unit propagation:
\begin{itemize}
\item if $c \in s$, then $c \in \UP{s}$;
\item if $\ell$ is a valid literal, then $\ell \in \UP{s}$;
\item if $c, \ell \in \UP{s}$ and $c'$ is the unit propagation of $c$ with $\ell$, then $c' \in \UP{s}$.
\end{itemize}
Furthermore $\Max{s}$ denotes the result of adding to $s$ all clauses that are valid or subsumed by some clause in $s$.
We further write $\Min{s}$ denote the setup obtained by removing from $s$ all valid clauses and all clauses subsumed by some other clause in $\VP{s}$.
To ease readability, we write $\WP{s}$ for $\Min{\UP{s}}$ and $\VP{s}$ for $\Max{\UP{s}}$.
A setup $s$ \emph{subsumes} a clause $c$ when some $c' \in s$ subsumes $c$.
\end{defi}

\begin{lem} \label{lem:setup-subsumption-up}
The following are equivalent:
\begin{enumeratethm}
\item $w \models c$ for all $c \in s$;
\item $w \models c$ for all $c \in \Min{s}$;
\item $w \models c$ for all $c \in \Max{s}$;
\item $w \models c$ for all $c \in \UP{s}$.
\end{enumeratethm}
\end{lem}

\begin{pf}
We show that (i) is equivalent to (ii), (iii), (iv).
The remaining equivalences then follow.

The only-if direction of (i) iff (ii) is trivial.
Conversely, suppose (ii) and $c \in s$ and $c \notin \Min{s}$.
Then $c$ is either valid or subsumed by some other clause.
In the first case, $w \models c$ by Lemma~\ref{lem:clause-valid}.
In the second case, there is some $c' \in \Min{s}$ that subsumes $c$ by Lemmas \ref{lem:clause-subsumption-transitive} and \ref{lem:clause-subsumption-asymmetric}.
By assumption, $w \models c'$, and by Lemma~\ref{lem:clause-subsumption} $w \models c$.

The if direction of (i) iff (iii) is trivial.
Conversely, suppose (i) and $c \in \Max{s}$ and $c \notin s$.
Then $c$ is either valid or subsumed by some other clause $c' \in s$.
In the first case, $w \models c$ by Lemma~\ref{lem:clause-valid}.
In the second case, $w \models c'$ by assumption, and by Lemma~\ref{lem:clause-subsumption} $w \models c$.

The if direction of (i) iff (iv) is trivial.
Conversely suppose (i) and $c \in \UP{s}$.
We show by induction on the length of the derivation of $c$ that $w \models c$.
\begin{itemize}
\item The base case for valid $\ell$ follows by Lemma~\ref{lem:literal-valid}.
\item The base case $c \in s$ is trivial.
\item For the induction step, let $c \in \UP{s}$ be the resolvent of $c', \ell \in \UP{s}$.
    By induction, $w \models c'$ and $w \models \ell$.
    By Lemma~\ref{lem:clause-unit-propagation}, $w \models c$.
    \qedhere
\end{itemize}
\end{pf}

\begin{defi}
Let $\UP[k]{s}$ be the set of clauses derivable by at most $k$ unit propagations from $s$ together with all valid literals.
\end{defi}

\begin{lem} \label{lem:setup-vp-subset-monotonic}
Let $s \subseteq s'$.
Then $\UP{s} \subseteq \UP{s'}$ and $\VP{s} \subseteq \VP{s'}$.
\end{lem}

\begin{pf}
We first show that if $c \in \UP{s}$ then $c \in \UP{s'}$ by induction on the length of the derivation of $c$.
\begin{itemize}
\item If $c \in \UP[0]{s}$, then $c$ is either valid or $c \in s \subseteq s'$, and so $c \in \UP[0]{s'}$.
\item If $c \in \UP[k+1]{s} \setminus \UP[k]{s}$, then $c$ is the unit propagation of $c', \ell \in \UP[k]{s}$.
    By induction, $c', \ell \in \UP[k]{s'}$, and so $c \in \UP[k+1]{s'}$.
\end{itemize}
Finally, if $c \in \VP{s}$, then $c$ is valid or subsumed by some $c' \in \UP{s} \subseteq \UP{s'}$, so $c \in \VP{s'}$.
\end{pf}

\begin{lem} \label{lem:setup-min-max}
\begin{enumeratethm}
\item $s \subseteq s'$ implies $\Max{s} \subseteq \Max{s'}$;
\item $\Min{s} \subseteq \Min{s'}$ implies $\Max{s} \subseteq \Max{s'}$;
\item $\Min{s} = \Min{s'}$ iff $\Max{s} = \Max{s'}$.
\end{enumeratethm}
\end{lem}

\begin{pf}
\begin{enumeratethm}
\item Suppose $s \subseteq s'$ and let $c \in \Max{s}$.
    If $c$ is valid, then $c \in \Max{s'}$.
    Otherwise $c$ is subsumed by some $c' \in s$.
    By assumption, $c' \in s'$, and so $c' \in \Max{s'}$.
    Hence $c \in \Max{s'}$.

\item Suppose $\Min{s} \subseteq \Min{s'}$ and let $c \in \Max{s}$.
    If $c$ is valid, then $c \in \Max{s'}$.
    Otherwise $c$ is subsumed by a clause in $s$, and by Lemmas \ref{lem:clause-subsumption-transitive} and \ref{lem:clause-subsumption-asymmetric}, $c$ is subsumed by some $c' \in \Min{s}$.
    By assumption, $c' \in \Min{s'}$, and so $c' \in \Max{s'}$.
    Hence also $c \in \Max{s'}$.

\item Suppose $\Max{s} = \Max{s'}$ and let $c \in \Min{s}$.
    Since $\Min{s} \subseteq \Max{s}$, we have $c \in \Max{s}$, and by assumption, $c \in \Max{s'}$.
    Moreover, $c$ is not valid and not subsumed by any clause in $s$ by assumption.
    Then $c$ is also not subsumed by any clause in $\Max{s} = \Max{s'}$, and in particular, not subsumed by any clause in $s'$.
    Hence $c \in \Min{s'}$.
    \qedhere
\end{enumeratethm}
\end{pf}

\begin{lem} \label{lem:setup-vp-wp-invariant}
Let $f(s) \in \{\VP{s},\WP{s}\}$.
\begin{enumeratethm}
\item $f(s \cup s') = f(\Min{s} \cup s')$;
\item $f(s \cup s') = f(\Max{s} \cup s')$;
\item $f(s \cup s') = f(\UP{s} \cup s')$.
\end{enumeratethm}
\end{lem}

\begin{pf}
\begin{enumeratethm}
\item First consider $\VP{s \cup s'} = \VP{\Min{s} \cup s'}$.
    The $\supseteq$ direction follows from Lemma~\ref{lem:setup-vp-subset-monotonic} since $s \cup s' \supseteq \Min{s} \cup s'$.
    Conversely, suppose $c \in \VP{s \cup s'}$.
    If $c$ is valid, then $c \in \VP{\Min{s} \cup s'}$.
    Otherwise $c$ is subsumed by some clause in $\UP{s \cup s'}$, and by Lemmas \ref{lem:clause-subsumption-transitive} and \ref{lem:clause-subsumption-asymmetric}, $c$ is subsumed by some $c' \UP{s \cup s'}$ which itself is not subsumed by any clause $\UP{s \cup s'}$.
    We show by induction on the length of the derivation of $c'$ that $c' \in \UP{\Min{s} \cup s'}$, which implies $c \in \VP{\Min{s} \cup s'}$.
    \begin{itemize}
    \item If $c' \in \UP[0]{\Min{s} \cup s'}$, then $c'$ is a valid literal or $c' \in \Min{s} \cup s$, and hence $c' \in \UP[0]{\Min{s} \cup s'}$.
    \item If $c' \in \UP[k+1]{s \cup s'} \setminus \UP[k]{s \cup s'}$, then it is the unit propagation of some $c'', \ell \in \UP[k]{s \cup s'}$ that are not subsumed by any clause in $\UP[k]{s \cup s'}$.
        For otherwise either the empty clause were in $\UP[k]{s \cup s'}$, or the unit propagation of the subsuming clauses would subsume $c$ by Lemma~\ref{lem:clause-subsumption-unit-propagation}, both of which would contradict the assumption.
        By induction, $c'', \ell \in \UP[k]{\Min{s} \cup s'}$, and so $c' \in \UP[k+1]{\Min{s} \cup s'}$.
    \end{itemize}

    The claim $\WP{s \cup s'} = \WP{\Min{s} \cup s'}$ follows by Lemma~\ref{lem:setup-min-max}.

\item Now consider $\WP{s \cup s'} = \WP{\Max{s} \cup s'}$.
    By (i), $\WP{\Max{s} \cup s'} = \WP{\Min{(\Max{s}) \cup s'}}$.
    Since $\Min{(\Max{s})} = \Min{s}$, we have $\WP{\Max{s} \cup s'} = \WP{\Min{s} \cup s'}$, and again by (i), $\WP{\Min{s} \cup s'} = \WP{s \cup s'}$.

    The claim $\VP{s \cup s'} = \VP{\Min{s} \cup s'}$ follows by Lemma~\ref{lem:setup-min-max}.

\item Now consider $\VP{s \cup s'} = \VP{\UP{s} \cup s'}$.
    The $\subseteq$ direction holds by Lemma~\ref{lem:setup-vp-subset-monotonic} since $s \cup s' \subseteq \UP{s} \cup s'$.
    Conversely, suppose $c \in \VP{\UP{s} \cup s'}$.
    If $c$ is valid, then $c \in \VP{s \cup s'}$.
    Otherwise $c$ is subsumed by some $c' \in \UP[k]{\UP[l]{s} \cup s'}$ for some $k, l$.
    We show by induction on $l$ and subinduction on $k$ that $c' \in \UP[k+l]{s \cup s'}$, which implies $c \in \VP{s \cup s'}$.
    \begin{itemize}
    \item For the base case let $l = 0$.
        Then $\UP[k]{\UP[0]{s} \cup s'} = \UP[k]{s \cup s'}$, so the claim holds.
    \item Now consider $l > 0$ and suppose $c' \in \UP[k]{\UP[l]{s} \cup s'}$.
        We show that $c' \in \UP[k+l]{s \cup s'}$ by subinduction on $k$.
        For $k = 0$, $\UP[0]{\UP[l]{s} \cup s'} \subseteq \UP[l]{s \cup s'}$, so the claim holds.
        For $k > 0$, suppose $c' \in \UP[k]{\UP[l]{s} \cup s'}$ and that $c'$ is the unit propagation of two clauses $c'', \ell \in \UP[k-1]{\UP[l]{s} \cup s'}$.
        By induction, $c'', \ell \in \UP[k-1+l]{s \cup s'}$.
        Hence $c \in \UP[k+l]{s \cup s'}$.
    \end{itemize}

    The claim $\WP{s \cup s'} = \WP{\Min{s} \cup s'}$ follows by Lemma~\ref{lem:setup-min-max}.
    \qedhere
\end{enumeratethm}
\end{pf}

\subsection{Formulas}

\begin{defi}[Formula]
The set of \emph{formulas} is the least set that contains all literals and all expressions $\neg \alpha$, $(\alpha \lor \beta)$, $\ex x \alpha$, $\K[k] \alpha$, $\M[k] \alpha$, $\OO \alpha$, $\G \alpha$ where $\alpha$ and $\beta$ are formulas, $x$ is a variable, and $k \geq 0$ is a natural number.
To denote the names or primitive terms that occur in a formula $\alpha$, we write $\Names(\alpha)$ and $\Terms(\alpha)$, respectively, optionally indexed with a term $t$ to restrict the set to terms of the same sort as $t$; and analogously for sets of formulas.
We use $\Lmod$ as a placeholder for $\K$ and $\M$.
\end{defi}

\begin{defi}[Length]
The length $|\alpha|$ of a formula $\alpha$ is defined as follows:
\begin{itemize}
\item $|t \e t'| = 3$;
\item $|(\alpha \lor \beta)| = 3 + |\alpha| + |\beta|$;
\item $|\neg \alpha| = 2^{m(\alpha)} + |\alpha|$;
\item $|\ex x \alpha| = 1 + |\alpha|$;
\item $|\Lmod[k] \alpha| = k + 1 + |\alpha|$;
\item $|\OO \phi| = 1 + |\phi|$;
\item $|\G \alpha| = 1 + |\alpha|$;
\end{itemize}
where $m(\alpha)$ is the number of modal operators in $\alpha$.
\end{defi}

The point of defining $|\neg \alpha| = 2^{m(\alpha)} + |\alpha|$ is that $|\neg \Lmod[k] \alpha| > |\Lmod[k] \neg \alpha|$ and likewise $|\neg \G \alpha| > |\G \neg \alpha|$.

\section{Only-Knowing} \label{app:o}

In this section we show Theorem~\ref{thm:o}, which says that limited only-knowing has a unique model (modulo unit propagation and subsumption).

\begin{lem} \label{lem:setup-fix-modelss}
Let $f(s) \in \{\UP{s}, \Min{s}, \Max{s}\}$.
Then $s_0,s \cup s',v \modelss \alpha$ iff $s_0,f(s) \cup s',v \cup \modelss \alpha$ iff $f(s_0),s \cup s',v \modelss \alpha$.
\end{lem}

\begin{pf}
By induction on $|\alpha|$.
\begin{itemize}
\item For the base case consider a clause $c$, which includes the base cases for $\ell$ and a clause $(\alpha \lor \beta)$.
    Then $s_0,s \cup s',v \modelss c$ iff $c \in \VP{s \cup s'}$ iff $c \in \VP{f(s) \cup s'}$ iff $s_0,f(s) \cup s',v \modelss c$.
    Moreover $s_0,s \cup s',v \modelss c$ iff $f(s_0),s \cup s',v \modelss c$.

\item The induction steps for $\neg (\alpha \lor \beta)$, $\ex x \alpha$, $\neg \ex x \alpha$, and $\neg \neg \alpha$ are trivial.

\item Now consider $\K[0] \alpha$.
    Firstly, $s_0,s \cup s',v \modelss \K[0] \alpha$ iff $s_0 \cup v$ is obviously inconsistent or $s_0,s_0 \cup v \modelss \alpha$ iff $s_0,f(s) \cup s',v \modelss \K[0]$.
    Secondly, $s_0,s \cup s',v \modelss \K[0] \alpha$ iff $s_0 \cup v$ is obviously inconsistent or $s_0,s_0 \cup v \modelss \alpha$ iff (by induction and by the fact that $\VP{s_0 \cup v} = \VP{f(s_0) \cup v}$) $f(s_0) \cup v$ is obviously inconsistent or $f(s_0),f(s_0) \cup v \modelss \alpha$ iff $f(s_0),s \cup s',v \modelss \K[0] \alpha$.

\item Now consider $\K[k+1] \alpha$.
    Firstly, $s_0,s \cup s',v \modelss \K[k+1] \alpha$ iff for some $t$ and all $n \in \Names_t$, $s_0,s \cup s',v \cup \{t \e n\} \modelss \K[k] \alpha$ iff (by induction) for some $t$ and all $n \in \Names_t$, $s_0,f(s) \cup s',v \cup \{t \e n\} \modelss \K[k] \alpha$ iff $s_0,f(s) \cup s',v \modelss \K[k+1] \alpha$.
    Analogously for the second claim.

\item Now consider $\M[0] \alpha$.
    Firstly, $s_0,s \cup s',v \modelss \M[0] \alpha$ iff $s_0 \cup v$ is not potentially inconsistent and $s_0,s_0 \cup v \modelss \alpha$ iff $s_0,f(s) \cup s',v \modelss \M[0]$.
    Secondly, $s_0,s \cup s',v \modelss \M[0] \alpha$ iff $s_0 \cup v$ is not potentially inconsistent $s_0,s_0 \cup v \modelss \alpha$ iff (by induction and by the fact that $\WP{s_0 \cup v} = \WP{f(s_0) \cup v}$) $f(s_0) \cup v$ is not potentially inconsistent $f(s_0),f(s_0) \cup v \modelss \alpha$ iff $f(s_0),s \cup s',v \modelss \M[0] \alpha$.

\item Now consider $\M[k+1] \alpha$.
    Firstly, $s_0,s \cup s',v \modelss \M[k+1] \alpha$ iff for some $t$ and $n \in \Names_t$, $s_0,s \cup s',v \cup \{t \e n\} \modelss \M[k] \alpha$ or $s_0,s \cup s',v \uplus_{s_0} (t \e n) \modelss \alpha$ iff (by induction) for some $t$ and all $n \in \Names_t$, $s_0,f(s) \cup s',v \cup \{t \e n\} \modelss \M[k] \alpha$ iff $s_0,f(s) \cup s',v \modelss \M[k+1] \alpha$.
    Secondly, $s_0,s \cup s, v \modelss \M[k+1] \alpha$ iff for some $t$ and $n \in \Names_t$, $s_0,s \cup s',v \cup \{t \e n\} \modelss \M[k] \alpha$ or $s_0,s \cup s',v \uplus_{s_0} (t \e n) \modelss \alpha$ iff (by induction and by the fact that $\VP{s_0 \cup v} = \VP{f(s) \cup v}$) for some $t$ and $n \in \Names_t$, $f(s_0),s \cup s',v \cup \{t \e n\} \modelss \M[k] \alpha$ or $f(s_0),s \cup s',v \uplus_{f(s_0)} (t \e n) \modelss \alpha$ iff $f(s_0),s \cup s',v \modelss \alpha$.

\item Firstly, $s_0,s \cup s',v \modelss \G \alpha$ iff $s_0|_{\Terms(\gnd(\alpha))},s \cup s',v \modelss \alpha$ iff (by induction) $s_0|_{\Terms(\gnd(\alpha))},f(s) \cup s',v \modelss \alpha$ iff $s_0,f(s) \cup s',v \modelss \alpha$.
    Secondly, $s_0,s \cup s',v \modelss \G \alpha$ iff $s_0|_{\Terms(\gnd(\alpha))},s \cup s',v \modelss \alpha$ iff (by induction and by the fact that $\WP{s_0} = \WP{f(s_0)}$) $f(s_0)|_{\Terms(\gnd(\alpha))},s \cup s',v \modelss \alpha$ iff $f(s_0),s \cup s',v \modelss \alpha$.

\item The induction steps for $\neg \K[k] \alpha$, $\neg \M[k] \alpha$, and $\neg \G \alpha$ are trivial.
    \qedhere
\end{itemize}
\end{pf}


\begin{lem} \label{lem:o-unique-vp}
Let $\phi$ be proper\plus.
Then $s_0 \modelss \OO \phi$ iff $\VP{s_0} = \VP{\gnd(\phi)}$.
\end{lem}

\begin{pf}
For the only-if direction suppose $s_0 \modelss \OO \phi$.
Then $s_0,s_0 \modelss \phi$.
Then $\gnd(\phi) \subseteq \VP{s_0}$, and hence clearly $\VP{\gnd(\phi)} \subseteq \VP{s_0}$.
Moreover, $s_0,\gnd(\phi) \modelss \phi$, and by assumption, there is no $\hat{s}_0$ with $\VP{\hat{s}_0} \subseteq \VP{s_0}$ and $s_0,\hat{s}_0 \modelss \phi$.
Thus $\VP{\gnd(\phi)} \supseteq \VP{s_0}$.
Together, this gives $\VP{s_0} = \VP{\gnd(\phi)}$.

For the if direction suppose $\VP{s_0} = \VP{\gnd(\phi)}$.
Then $c \in \VP{s_0}$ for every $c \in \gnd(\phi)$, so $s_0,s_0 \modelss c$, and thus $s_0,s_0 \modelss \phi$.
Suppose $\hat{s}_0$ is such that $\VP{\hat{s}_0} \subsetneq \VP{s_0}$.
Let $c \in \VP{s_0}$ but $c \notin \VP{\hat{s}_0}$.
Then $c$ is not valid.
If $c$ is subsumed by some other clause $c' \in \VP{s_0}$, then $c' \notin \VP{\hat{s}_0}$; hence we can assume that $c$ is not subsumed by any other clause in $\VP{s_0}$.
If $c$ is the unit propagation of two clauses $c', \ell \in \VP{s_0}$, then $c', \ell \notin \VP{\hat{s}_0}$; hence we can assume that $c$ is not the unit propagation of any other clauses.
By these assumptions and since $\VP{s_0} = \VP{\gnd(\phi)}$, we have $c \in \gnd(\phi)$ but $c \notin \VP{\hat{s}_0}$.
Hence $s_0,\hat{s}_0 \not\modelss \phi$.
\end{pf}

\begin{thm} \label{thm:o}
Let $\phi$ be proper\plus.
$\OO \phi \entailss \alpha$ iff $\gnd(\phi) \modelss \alpha$.
\end{thm}

\begin{pf}
For the only-if direction, suppose $\OO \phi \entailss \alpha$.
By Lemma~\ref{lem:o-unique-vp}, $\gnd(\phi) \modelss \OO \phi$, and hence $\gnd(\phi) \modelss \alpha$.
Conversely, suppose $\gnd(\phi) \modelss \alpha$ and $s_0 \modelss \OO \phi$.
By Lemma~\ref{lem:o-unique-vp}, $\VP{s_0} = \VP{\gnd(\phi)}$, and by Lemma~\ref{lem:setup-fix-modelss}, $s_0 \modelss \alpha$.
\end{pf}

\section{Soundness} \label{app:soundness}

In this section we prove Theorem~\ref{thm:soundness} that states the soundness of limited belief.

%

\begin{defi}
We denote by $\sigma_\L$ the result of replacing in $\sigma$ every $\Lmod[k]$ with $\Lmod$.
We abbreviate $[\phi] = \{w \mid w \models \phi\}$, and $[\Phi] = \bigcap_{\phi \in \Phi} [\phi]$.
\end{defi}

\begin{lem} \label{lem:setup-obviously-inconsistent}
Suppose $s$ is obviously inconsistent.
Then there is no $w$ such that $w \models c$ for all $c \in s$.
\end{lem}

\begin{pf}
Suppose $w \models c$ for all $c \in s$.
By Lemma~\ref{lem:setup-subsumption-up}, $w \models c$ for all $c \in \Max{s}$.
By assumption, $n \n n \in \Max{s}$.
Thus $w \models n \n n$.
Contradiction.
\end{pf}

\begin{lem} \label{lem:setup-not-potentially-inconsistent}
Suppose $s$ is not potentially inconsistent.
Then there is a $w$ such that $w \models c$ for all $c \in s$.
\end{lem}

\begin{pf}
Let $L = \{\ell \mid \ell \in c \in \WP{s}\}$.
By assumption, $L$ contains no two complementary literals and for no $t$ all $t \e n$ for $n \in \Names_t$ occur in $\WP{s}$.
Let $w$ be such that $w(t) = n$ for every $t \e n \in L$.
Such $w$ is well-defined because by assumption, if $t \e n \in L$, then $n \in \Names_t$ for otherwise it would contain complementary literals, and neither $t \n n \in L$ nor $t \e n' \in L$ for $n'$ distinct from $n$.
Moreover for every $t \n n \in L$ there is an $n' \in \Names_t$ such that $t \n n' \notin L$, so we can let $w(t) = n'$.
By assumption, $s$ is not obviously inconsistent, so every clause in $s$ is subsumed by a literal in $L$.
Hence by construction $w \models c$ for every $c \in s$.
\end{pf}

\begin{lem} \label{lem:k-m-setup-two-ignored}
For all $s_1, s_2$, $s_0,s_1,v \modelss \Lmod[k] \alpha$ iff $s_0,s_2,v \modelss \Lmod[k] \alpha$.
\end{lem}

\begin{pf}
By induction on $k$.
Base case and induction step are obvious as $s_1, s_2$ are not mentioned on the right-hand sides of the semantic rules.
\end{pf}

\begin{lem} \label{lem:m-consistent}
Suppose $s_0,s,v \modelss \M[k] \alpha$.
Then there is a $w$ such that $w \models c$ for all $c \in s_0$.
\end{lem}

\begin{pf}
By induction on $k$.
\begin{itemize}
\item Base case $k = 0$: since $s_0,s,v \modelss \M[0] \alpha$ requires $s_0 \cup v$ to be not potentially inconsistent, which by Lemma~\ref{lem:setup-not-potentially-inconsistent} gives the lemma.
\item The induction step is trivial.
    \qedhere
\end{itemize}
\end{pf}

\begin{lem} \label{lem:soundness-s0}
Let $\phi$ be proper\plus, $\alpha$ be without $\OO, \G$, and without negated $\Lmod[k]$, and $[s_0] = \gnd(\phi)$.
For every $\VP{s} \supseteq \VP{s_0}$, if $s_0,s \modelss \alpha$, then $[s_0],w \models \alpha$ for all $w \in [s]$.
\end{lem}

\begin{pf}
We first prove the lemma by induction on $|\alpha|$.
\begin{itemize}
\item For the base case consider a clause $c$, which includes the base cases for $\ell$ and a clause $(\alpha \lor \beta)$.
    Suppose $s_0,s \modelss c$.
    Then $c \in \VP{s}$.
    Then for all $w$ with $w \models c'$ for all $c' \in \VP{s}$ also $w \models c$.
    By Lemma~\ref{lem:setup-subsumption-up}, for every $w$ with $w \models c'$ for all $c' \in s$ also $w \models c$.
    Thus $e,w \models c$ for all $w \in [s]$.

\item The induction steps for $\neg (\alpha \lor \beta)$, $\ex x \alpha$, $\neg \ex x \alpha$, and $\neg \neg \alpha$ are trivial.

\item Consider $\K[k] \alpha$.
    We define $e_0,e \models \K \alpha_\L$ iff $e_0,w \models \K \alpha_\L$ for all $w \in e$.
    We show by subinduction on $k$ that $s_0,s,v \modelss \K[k] \alpha$ implies $[s_0],[s_0 \cup v] \models \K \alpha_\L$.
    Since $[s_0] = [s_0 \cup \emptyset]$, we obtain that $s_0,s \modelss \K[k] \alpha$ implies $[s_0] \models \K \alpha_\L$.

    Consider $k = 0$ and suppose $s_0,s,v \modelss \K[0] \alpha$.
    If $s \cup v$ is obviously inconsistent, then by Lemma~\ref{lem:setup-obviously-inconsistent}, $[s \cup v] = \emptyset$ and hence trivially $[s_0],[s_0 \cup v] \models \K \alpha_\L$.
    If $s_0,s_0 \cup v \modelss \alpha$, then by induction $[s_0],w \modelss \alpha_\L$ for all $w \in [s_0 \cup v]$.
    Then $[s_0],[s_0 \cup v] \models \K \alpha_\L$.

    Consider $k + 1$ and suppose $s_0,s,v \modelss \K[k+1] \alpha$.
    For some $t \in \Terms$ and all $n \in \Names_t$, $s_0,s,v \cup \{t \e n\} \modelss \K[k+1] \alpha$.
    By subinduction, for all $n \in \Names_t$, $[s_0],[s_0 \cup v \cup \{t \e n\}] \models \K \alpha_\L$.
    Since $[s_0 \cup v] = [s_0 \cup v \cup \{t \e n\}]$, $[s_0],[s_0 \cup v] \modelss \K \alpha_\L$.

\item Consider $\M[k] \alpha$.
    We show by subinduction on $k$ that $s_0,s,v \modelss \M[k] \alpha$ implies $[s_0] \modelss \M \alpha_\L$.

    Consider $k = 0$ and suppose $s_0,s,v \modelss \M[0] \alpha$.
    By assumption, $s_0 \cup v$ is not potentially inconsistent, and $s_0,s_0 \cup v \modelss \alpha$.
    Then by Lemma~\ref{lem:setup-not-potentially-inconsistent}, $[s_0 \cup v] \neq \emptyset$, and by induction, $[s_0],w \models \alpha_\L$ for all $w \in [s_0 \cup v]$.
    In particular, there is $w \in [s_0 \cup v]$ such that $[s_0],w \models \alpha_\L$.
    Thus the same $w \in [s_0]$ satisfies $[s_0],w \models \alpha_\L$.
    Thus $[s_0] \models \M \alpha_\L$.

    Consider $k + 1$ and suppose $s_0,s,v \modelss \M[k+1] \alpha$.
    Then for some $t \e n$, either $s_0,s,v \cup \{t \e n\} \modelss \M[k] \alpha$ or $s_0,s,v \uplus_s0 (t \e n) \modelss \M[k] \alpha$.
    In either case by subinduction $[s_0] \models \M \alpha_\L$.
    \qedhere
\end{itemize}
\end{pf}

\begin{lem} \label{lem:soundness}
Let $\phi$ be proper\plus, $\sigma$ be subjective, without $\OO, \G$, and without negated $\Lmod[k]$, and $[s_0] = \gnd(\phi)$.
If $s_0,\emptyset \modelss \sigma$, then $[s_0] \models \sigma$.
\end{lem}

\begin{pf}
By induction on $|\sigma|$.
\begin{itemize}
\item For the base case consider a clause $c$, which includes the base cases for $\ell$ and a clause $(\alpha \lor \beta)$.
    By assumption, $c \in \VP{\emptyset}$, so $c$ is valid, and by Lemma~\ref{lem:clause-valid}, $[s_0] \models c$.

\item The induction steps for $\neg (\alpha \lor \beta)$, $\ex x \alpha$, $\neg \ex x \alpha$, and $\neg \neg \alpha$ are trivial.

\item Now consider $\K[0] \alpha$.
    If $s_0,\emptyset \modelss \K[k] \alpha$, then by Lemma~\ref{lem:k-m-setup-two-ignored}, $s_0,[s_0],\emptyset \modelss \K[k] \alpha$, and by the Lemma~\ref{lem:soundness-s0}, $[s_0],w \models \K \alpha_\L$ for all $w \in [s_0]$, and so $[s_0] \models \K \K \alpha_\L$ and thus $[s_0] \modelss \K \alpha_\L$.

\item Now consider $\M[0] \alpha$.
    If $s_0,\emptyset \modelss \M[k] \alpha$, then by Lemma~\ref{lem:k-m-setup-two-ignored}, $s_0,[s_0],\emptyset \modelss \M[k] \alpha$, and by the Lemma~\ref{lem:soundness-s0}, $[s_0],w \models \M \alpha_\L$ for all $w \in [s_0]$, and since $[s_0] \neq \emptyset$ by Lemma~\ref{lem:m-consistent}, $[s_0] \models \M \alpha_\L$.
    \qedhere
\end{itemize}
\end{pf}

\begin{repthm}[Soundness]{thm:soundness}
Let $\phi$ be proper\plus, $\sigma$ be subjective, without $\OO, \G$, and without negated $\Lmod[k]$.
If $\OO \phi \entailss \sigma$, then $\OO \phi \entails \sigma_\L$.
\end{repthm}

\begin{pf}
Suppose $\OO \phi \entailss \sigma$ and $e \models \OO \phi$.
Let $s_0 = \gnd(\phi)$.
Then $[s_0]$ is the unique set of worlds such that $[s_0] \models \OO \phi$, so we need to show that $[s_0] \models \sigma_\L$.
By Theorem~\ref{thm:o}, $s_0 \modelss \OO \phi$, and by assumption, $s_0 \modelss \sigma$.
Then $s_0,\emptyset \modelss \sigma$, and by Lemma~\ref{lem:soundness}, $[s_0] \models \sigma_\L$.
\end{pf}

\section{Eventual Completeness} \label{app:eventual-completeness}

Here we prove the eventual completeness result for propositional limited belief, Theorem~\ref{thm:eventual-completeness}.

\begin{defi}
A formula without quantifiers is called \emph{propositional}.
\end{defi}

\begin{defi}
Let $s(w,T) = s \cup \{t \e n \mid w(t) = n \text{ and } t \in T\}$.
\end{defi}

\begin{lem} \label{lem:k-monotonic}
$\Lmod[k] \alpha \entailss \Lmod[k+1] \alpha$.
\end{lem}

\begin{pf}
By induction on $k$.
Suppose $s_0 \modelss \Lmod[k] \alpha$.
\begin{itemize}
\item If $s_0,s,v \modelss \K[0] \alpha$, then $s_0$ is obviously inconsistent or $s_0,s_0 \cup v,\emptyset \modelss \alpha$.
    Let $t \notin \Terms(s_0 \cup v \cup \{\alpha\})$ be a primitive term and $n \in \Terms_t$ be arbitrary.
    If $s_0$ is obviously inconsistent, then clearly $s_0 \cup \{t \e n\}$ is obviously inconsistent.
    If $s_0,s_0 \cup v,\emptyset \modelss \alpha$, then $s_0,s_0 \cup v \cup \{t \e n\},\emptyset \modelss \alpha$, as can be shown by a trivial induction on $|\alpha|$.
    In either case, $s_0,s,v \cup \{t \e n\} \modelss \K[0] \alpha$ for every $n \in \Terms_t$.
    Then $s_0,s,v \modelss \K[1] \alpha$.

\item If $s_0,s,v \modelss \M[0] \alpha$, then $s_0$ is not potentially inconsistent and $s_0,s_0 \cup v,\emptyset \modelss \alpha$.
    Let $t \notin \Terms(s_0 \cup v \cup \{\alpha\})$ be a primitive term and $n \in \Terms_t$ be arbitrary.
    Then clearly $s_0 \cup \{t \e n\}$ is not potentially inconsistent, and as in the case for $\K[0]$, $s_0,s_0 \cup v \cup \{t \e n\},\emptyset \modelss \alpha$.
    In either case, $s_0,s,v \cup \{t \e n\} \modelss \M[0] \alpha$.
    Then $s_0,s,v \modelss \M[1] \alpha$.

\item The induction step is trivial.
    \qedhere
\end{itemize}
\end{pf}

\begin{lem} \label{lem:setup-monotonic}
Let $s_0,s \modelss \alpha$ and $s' \supseteq s$.
Then $s_0,s' \modelss \alpha$.
\end{lem}

\begin{pf}
By induction on $|\alpha|$.
\begin{itemize}
\item For every clause $c$ with $s_0,s \modelss c$, we have $c \in \VP{s} \subseteq \VP{s'}$, and thus $s_0,s' \modelss c$.
\item The induction steps are trivial.
    \qedhere
\end{itemize}
\end{pf}

\begin{lem} \label{lem:setup-all-combinations}
Let $s_0$ be finite, $\alpha$ be propositional, $s \supseteq s_0$, $T \supseteq \Terms(s_0 \cup \{\alpha\})$, and $T' = \{t_1,\ldots,t_j\} \subseteq T$ be maximal such that for every $n \in \Names_t$, $t \e n \notin s_0$.
If $s_0,s(w,T) \modelss \alpha$ for all $w \in [s_0]$, then $s_0,s \cup \{t_1 \e n_1,\ldots,t_j \e n_j\} \modelss \alpha$ for all $n_i \in \Names_{t_i}$.
\end{lem}

\begin{pf}
By induction on $j$.
Suppose $s_0,s(w,T) \models \alpha_\L$ for all $w \in [s_0]$.
\begin{itemize}
\item Suppose $j = 0$.

    If there is some $w \in [s_0]$, then $s(w,T) = s$, for otherwise there is a $t \in T \setminus T'$ such that $t \e n \in s$ and $w(t) \neq n$ and thus $w \not\models t \e n$, which contradicts the assumption.
    The claim follows trivially $s(w,T) = s$.

    Otherwise, for every $w$ there must be some $c \in s_0 \subseteq s$ such that $w \not\models c$, which implies $w \models \neg \ell$ for every $\ell \in c$.
    This implies that such $\ell$ is not valid by Lemma~\ref{lem:literal-valid}, and since $\ell$ mentions a name on the left-hand side by assumption, $\entails \neg \ell$ by Lemma~\ref{lem:literal-names-not-valid}, so $\ell$ is of the form $n \n n$ or $n \e n'$ for two distinct names $n, n'$.
    As $n \e n, n \n n' \in \UP{s}$ since they are valid literals, $\UP{s}$ contains the empty clause.
    A simple subinduction on $|\alpha|$ then shows that $s_0,s \modelss \alpha$.

\item Now consider $j+1$ and suppose the claim holds for $\{t_1,\ldots,t_j\}$.
    Suppose $s_0,(s \cup \{{t_{j+1} = n_{j+1}}\}(w,T) \modelss \alpha$ for all $w \in [s_0]$.
    By Lemma~\ref{lem:setup-monotonic}, $s_0,(s \cup \{{t_{j+1} = n_{j+1}}\})(w,T) \models \alpha_\L$ for every $n_{j+1} \in \Names_{t_{j+1}}$ and all $w \in [s_0]$.
    By induction, $s_0,s \cup \{t_1 \e n_1, \ldots, {t_{j+1} = n_{j+1}}\} \modelss \alpha$.
    \qedhere
\end{itemize}
\end{pf}

\begin{lem} \label{lem:setup-all-combinations-inconsistent}
Let $s_0$ be finite, $T \supseteq \Terms(s_0 \cup \{\alpha\})$, and $T' = \{t_1,\ldots,t_j\} \subseteq T$ be maximal such that for every $n \in \Names_t$, $t \e n \notin s_0$.
If $[s_0] = \emptyset$, then $s_0 \cup \{t_1 \e n_1,\ldots,t_j \e n_j\}$ is obviously inconsistent.
\end{lem}

\begin{pf}
Suppose $[s_0] = \emptyset$.
By Lemma~\ref{lem:setup-all-combinations}, $s_0 \cup \{t_1 \e n_1,\ldots,t_j \e n_j\} \modelss n \n n$ for arbitrary $n$.
Then $\VP{s_0 \cup \{t_1 \e n_1,\ldots,t_j \e n_j\}}$ must contain the empty clause and therefore it is obviously inconsistent.
\end{pf}

\begin{lem} \label{lem:setup-world-not-potentially-inconsistent}
Let $s_0$ be finite, $w \in [s_0]$, and $T \supseteq \Terms(s_0)$, $\alpha$ be propositional.
Then $s_0(w,T)$ is not potentially inconsistent.
\end{lem}

\begin{pf}
By assumption, $w \models c$ for all $c \in s_0$.
Furthermore, $w \models t \e n$ for every primitive term $t$, in particular for $t \in T$, and name $n$ identical to $w(t)$.
Thus $w \models c$ for all $c \in s_0(w,T)$.
By Lemma~\ref{lem:setup-obviously-inconsistent}, $s_0(w,T)$ is not obviously inconsistent

By Lemma~\ref{lem:setup-subsumption-up}, $w \models c$ for every $c \in \UP{s_0(w,T)}$.
Thus for every clause $c \in \UP{s_0(w,T)}$, there is some $\ell \in c$ such that $w \models \ell$.
If $\ell$ is of the form $t \e n$, then $w(t) = n$ and hence $t \e n \in \UP{s_0(w,T)}$.
If $\ell$ is of the form $t \n n'$, then $w(t) = n$ for some distinct $n$ and hence $t \e n \in \UP{s_0(w,T)}$.
In either case, $t \e n$ subsumes $c$, and so either $c$ is just the unit clause $t \e n$ or $c \notin \WP{s_0(w,T)}$.
Since $s_0(w,T)$ is not obviously inconsistent, there is at most one one $n$ per $t$ such that $t \e n \in \WP{s_0(w,T)}$.
Thus $s_0(w,T)$ is not potentially inconsistent.
\end{pf}

\begin{lem} \label{lem:completeness}
Let $s_0$ be finite, $\alpha_k$ be propositional with $l \geq k$ for every $\Lmod[l]$ it mentions, and $T \supseteq \Terms(s_0 \cup \{\alpha_k\})$.
If $w \in [s_0]$ with $[s_0],w \models \alpha_\L$, then $s_0,s_0(w,T) \modelss \alpha_{|T|}$.
\end{lem}

\begin{pf}
By induction on $|\alpha_k|$.
\begin{itemize}
\item Suppose $w \in [s_0]$ and $[s_0],w \models c$.
    Then $w \models \ell$ for some $\ell \in c$.
    If $\ell$ has a function as left-hand side, then $\ell \in s_0(w,T)$.
    Otherwise, $\entails \ell$, and by Lemma~\ref{lem:literal-valid}, $\ell \in \UP{s_0(w,T)}$.
    Thus $s_0,s_0(w,T) \modelss c$.

\item The induction steps for $\neg (\alpha \lor \beta)$, $\ex x \alpha$, $\neg \ex x \alpha$, and $\neg \neg \alpha$ are trivial.

\item Suppose $w \in [s_0]$ and $[s_0],w \models \K \alpha_\L$.
    Then $[s_0] \neq \emptyset$.
    Then $[s_0],w' \models \alpha_\L$ for all $w' \in [s_0]$.
    By induction, $s_0,s_0(w',T) \modelss \alpha_{|T|}$ for all $w' \in [s_0]$.
    By Lemma~\ref{lem:setup-all-combinations}, $s_0,s_0 \cup \{t_1 \e n_1, \ldots, t_j \e n_j\} \modelss \alpha_{|T|}$ for all $n_i \in \Names_{t_i}$.
    Thus $s_0 \modelss \K_j \alpha_{|T|}$.
    By Lemma~\ref{lem:k-monotonic}, $s_0 \modelss \K[l] \alpha_{|T|}$ for arbitrary $l \geq |T|$.

\item Suppose $w \in [s_0]$ and $[s_0],w \models \neg \K \alpha_\L$.
    Then $[s_0],w \not\models \K \alpha_\L$.
    By Theorem~\ref{thm:soundness}, $s_0 \not\modelss \K[l] \alpha$ for arbitrary $l$.
    Thus $s_0 \modelss \neg \K[l] \alpha$.

\item Suppose $w \in [s_0]$ and $[s_0],w \models \M \alpha_\L$.
    Then $[s_0] \neq \emptyset$.
    Then $[s_0],w' \models \alpha_\L$ for some $w' \in [s_0]$.
    By induction, $s_0,s_0(w',T) \modelss \alpha_{|T|}$ for this $w'$.
    By Lemma~\ref{lem:setup-world-not-potentially-inconsistent}, $s_0(w',T)$ is not potentially inconsistent.
    Thus and since $s_0(w',T)$ is the result of adding $|T|$ equality literals to $s_0$, we have $s_0 \modelss \M_{|T|} \alpha$.
    By Lemma~\ref{lem:k-monotonic}, $s_0 \modelss \M[l] \alpha_{|T|}$ for arbitrary $l \geq |T|$.

\item Suppose $w \in [s_0]$ and $[s_0],w \models \neg \M \alpha_\L$.
    Then $[s_0],w \not\models \M \alpha_\L$.
    By Theorem~\ref{thm:soundness}, $s_0 \not\modelss \M[l] \alpha$ for arbitrary $l$.
    Thus $s_0 \modelss \neg \M[l] \alpha$.
    \qedhere
\end{itemize}
\end{pf}

\begin{lem} \label{lem:k-m-exclusive}
\begin{enumeratethm}
\item If $s_0 \modelss \K[k] \alpha$, then $s_0 \not\modelss \M[l] \neg \alpha$.
\item If $s_0 \modelss \M[k] \alpha$, then $s_0 \not\modelss \K[l] \neg \alpha$.
\end{enumeratethm}
\end{lem}

\begin{pf}
\begin{enumeratethm}
\item Suppose $s_0 \modelss \K[k] \alpha$.
    Then $[s_0] \models \K \alpha_\L$ by Theorem~\ref{thm:soundness}.
    Then $[s_0] \not\models \M \neg \alpha_\L$.
    Then $s_0 \not\modelss \M[l] \neg \alpha$ by Theorem~\ref{thm:soundness}.

\item Suppose $s_0 \modelss \M[k] \alpha$.
    Then $[s_0] \models \M \alpha_\L$ by Theorem~\ref{thm:soundness}.
    Then $[s_0] \not\models \K \neg \alpha_\L$.
    Then $s_0 \not\modelss \K[l] \neg \alpha$ by Theorem~\ref{thm:soundness}.
    \qedhere
\end{enumeratethm}
\end{pf}

\begin{repthm}[Eventual completeness]{thm:eventual-completeness}
Let $\phi, \sigma_l$ be propositional, $\phi$ be proper\plus, $\sigma_l$ be subjective, without $\OO, \G$, and $k \geq l$ for all $\Lmod[k]$ in $\sigma_l$.
Then $\OO \phi \entails \sigma_\L$ implies that there is a $k$ such that $\OO \phi \entailss \sigma_k$.
\end{repthm}

\begin{pf}
Let $s_0 = \gnd(\phi)$, which is the unique (modulo $\VP{}$) model of $\OO \phi$, and $[s_0] \models \OO \phi$.
Let $k = |\Terms(s_0 \cup \{\sigma\})|$.
The proof is by induction on $|\sigma|$.
\begin{itemize}
\item For the base case suppose $[s_0] \models c$ for a clause $c$, which includes the base cases for $\ell$ and a clause $(\alpha \lor \beta)$.
    Then $c$ is valid by Lemma~\ref{lem:clause-valid}, and hence $c \in \VP{s}$, and so $s_0,s \modelss c$.

\item The induction steps for $\neg (\alpha \lor \beta)$, $\ex x \alpha$, $\neg \ex x \alpha$, and $\neg \neg \alpha$ are trivial.

\item Suppose $[s_0] \models \K \alpha_\L$.
    If $[s_0] \neq \emptyset$, then by Lemma~\ref{lem:completeness} $s_0 \modelss \K[l] \alpha$ for arbitrary $l \geq k$.
    If $[s_0] = \emptyset$, then by Lemma~\ref{lem:setup-all-combinations-inconsistent} there are $t_1,\ldots,t_k$ terms such that $s_0 \cup \{t_1 \e n_1, \ldots, t_k \e n_k\}$ is obviously inconsistent for all $n_i \in \Names_{t_i}$, and hence $s_0 \modelss \K[l] \alpha$ for arbitrary $l \geq k$.

\item Suppose $[s_0] \models \neg \K \alpha_\L$.
    Then $[s_0] \models \M \neg \alpha_\L$.
    Then $s_0 \modelss \M[l] \neg \alpha$ for arbitrary $l \geq k$.
    Then $s_0 \not\modelss \K[l] \neg \neg \alpha$ for arbitrary $l \geq k$ by Lemma~\ref{lem:k-m-exclusive}.
    Then $s_0 \modelss \neg \K[l] \alpha$ for arbitrary $l \geq k$.

\item Suppose $[s_0] \models \M \alpha_\L$.
    Then $[s_0] \neq \emptyset$, and by Lemma~\ref{lem:completeness}, $s_0 \modelss \M[l] \alpha$ for arbitrary $l \geq k$.

\item Suppose $[s_0] \models \neg \M \alpha_\L$.
    Then $[s_0] \models \K \neg \alpha_\L$.
    Then $s_0 \modelss \K[l] \neg \alpha$ for arbitrary $l \geq k$.
    Then $s_0 \not\modelss \M[l] \neg \neg \alpha$ for arbitrary $l \geq k$ by Lemma~\ref{lem:k-m-exclusive}.
    Then $s_0 \modelss \neg \M[l] \alpha$ for arbitrary $l \geq k$.
    \qedhere
\end{itemize}
\end{pf}

\section{Representation Theorem} \label{app:representation}

This section proves the limited version of Levesque's representation theorem, Theorem~\ref{thm:representation}.
A key lemma on the way is Lemma~\ref{lem:res} about the $\RES{}{}$ operator.

\subsection{The $\RES{}{}$ Operator}

\begin{defi}
Let $\phi$ be proper\plus, $T$ be a set of primitive terms, and $\psi$ be objective.
Then $\RES{\phi,T}{\Lmod[k] \psi}$ is defined as follows:
\begin{itemize}
\item if $\psi$ mentions a free variable $x$:\\
    $\bigvee_{n \in \Names_x(\phi) \cup \Names_x(\psi)} (x \e n \land \RES{\phi,T}{\Lmod[k] \psi^x_n}) \lor{}$\\
    $\big(\bigwedge_{n \in \Names_x(\phi) \cup \Names_x(\psi)} x \n n \land \RES{\phi,T}{\Lmod[k] \psi^x_{\hat{n}}}^{\hat{n}}_x\big)$\\
    where $\hat{n} \in \Names_x \setminus (\Names_x(\phi) \cup \Names_x(\psi))$ is a some new name;
\item if $\psi$ mentions no free variables:\\
    $\TRUE$ if $\gnd(\phi)|_T \modelss \Lmod[k] \psi$, and $\neg \TRUE$ otherwise,\\
    where $\TRUE$ stands for $\ex x x \e x$.
\end{itemize}
\end{defi}

\begin{defi}
A name involution is a sort-preserving bijection $\ast : \Names \rightarrow \Names$ such that $n^\ast{}^\ast \e n$.
\end{defi}

\begin{lem} \label{lem:setup-disjoint-up}
Suppose $\Terms(s_0)$ and $\Terms(s_1)$ are disjoint, and $c \in \UP{s_0 \cup s_1}$ mentions terms from $\Terms(s_i)$ or no terms at all.
Then $c \in \UP{s_i}$ or $s_{1-i}$ is obviously inconsistent.
\end{lem}

\begin{pf}
By induction on the length of the derivation of $c$.
\begin{itemize}
\item For the base case, if $c \in s_0 \cup s_1$ mentions a term from $\Terms(s_i)$, then $c \in \UP{s_i}$.
\item For the induction step, suppose $c$ is the unit propagation of $c', \ell \in \UP{s_0 \cup s_1}$.

    First suppose $\ell$ is valid and $c'$ mentions terms from $\Terms(s_i)$.
    By induction, $c' \in \UP{s_i}$.
    Then $c \in \UP{s_i}$.
    Since $\Terms(c) \subseteq \Terms(c')$ and if $c$ is the empty clause then $s_i$ is obviously inconsistent, the lemma follows.

    Now suppose $\ell$ is invalid.
    By induction, $\ell \in \UP{s_i}$ or $s_{1-i}$ is obviously inconsistent.
    Then $s_0$ or $s_1$ are obviously inconsistent.

    Finally suppose $\ell$ is neither valid nor invalid.
    Then $\ell$ must mention a term, and $c'$ must also mention that term.
    By induction, $c', \ell \in \UP{s_i}$.
    Then $c \in \UP{s_i}$.
    Since $\Terms(c) \subseteq \Terms(c')$ and if $c$ is the empty clause then $s_i$ is obviously inconsistent, the lemma follows.
    \qedhere
\end{itemize}
\end{pf}

\begin{lem} \label{lem:setup-disjoint-vp}
Suppose $\Terms(c) \cup \Terms(s_1)$ and $\Terms(s_2)$ are disjoint.
Then $c \in \VP{s_1 \cup s_2}$ iff $c \in \VP{s_1}$ or $s_2$ is potentially inconsistent.
\end{lem}

\begin{pf}
The if direction is trivial.
Conversely, suppose $c \in \VP{s_1 \cup s_2}$.
Then $c$ is subsumed by some $c' \in \UP{s_1 \cup s_2}$.
Then $\Terms(c') \subseteq \Terms(c)$.
By Lemma~\ref{lem:setup-disjoint-up}, $c' \in \UP{s_1}$ or $s_2$ is obviously inconsistent, and so the lemma holds.
\end{pf}

\begin{lem} \label{lem:setup-involution}
Let $f(s) \in \{\UP{s}, \Min{s}, \Max{s}\}$ and $\ast$ be a name involution.
Then $f(s)^\ast = f(s^\ast)$.
\end{lem}

\begin{pf}
\begin{itemizethm}
\item We first show $\UP{s}^\ast = \UP{s^\ast}$.

    For the $\subseteq$ direction suppose $c^\ast \in \UP{s}^\ast$.
    Then $c \in \UP{s}$.
    If $c \in s$ or $c$ is valid, then $c^\ast \in s^\ast$ or $c^\ast$ is valid, and hence $c^\ast \in \UP{s^\ast}$.
    Otherwise $c$ is the unit propagation of $c', \ell \in \UP{s}$, and then a trivial induction on the length of the derivation shows that then $c'{}^\ast, \ell^\ast \in \UP{s^\ast}$ and then $c^\ast \in \UP{s^\ast}$.

    For the $\supseteq$ direction suppose $c^\ast \in \UP{s^\ast}$.
    If $c^\ast \in s^\ast$ or $c^\ast$ is valid, then $c \in s$ or $c$ is valid, and then $c \in \UP{s}$ and thus $c^\ast \in \UP{s}^\ast$.
    Otherwise $c$ is the unit propagation of $c'{}^\ast, \ell^\ast \in \UP{s^\ast}$, and then a trivial induction on the length of the derivation shows that $c'{}^\ast, \ell^\ast \in \UP{s}^\ast$, and then $c', \ell \in \UP{s}$, and so $c \in \UP{s}$ and $c^\ast \in \UP{s}^\ast$.

\item Now we show that $(\Min{s})^\ast = \Min{(s^\ast)}$.
    We have $c^\ast \in (\Min{s})^\ast$ iff $c \in \Min{s}$ iff no other clause in $s$ subsumes $c$ iff no other clause in $s^\ast$ subsumes $c^\ast$ iff $c^\ast \in \Min{(s^\ast)}$.

\item Finally we show that $(\Max{s})^\ast = \Max{(s^\ast)}$.
    We have $c^\ast \in (\Max{s})^\ast$ iff $c \in \Max{s}$ iff some clause in $s$ subsumes $c$ iff some clause in $s^\ast$ subsumes $c^\ast$ iff $c^\ast \in \Max{(s^\ast)}$.
    \qedhere
\end{itemizethm}
\end{pf}

\begin{lem} \label{lem:objective-involution}
Let $\psi$ be objective, $\ast$ be a name involution and $\Terms(s'_0)$ be disjoint from $\Terms(s_0) \cup \Terms(\gnd(\psi))$.
Then $\cdot,s_0 \cup s'_0 \modelss \psi$ iff $\cdot,s_0^\ast \cup s'_0 \modelss \psi^\ast$.
\end{lem}

\begin{pf}
By induction on $|\psi|$.
\begin{itemize}
\item For the base case consider a clause $c$.
    By assumption, $\Terms(c) \subseteq \Terms(\gnd(\psi))$.
    Then $\cdot,s_0 \cup s'_0 \modelss c$ iff $c \in \VP{s_0 \cup s'_0}$ iff (by Lemma~\ref{lem:setup-disjoint-vp}) $c \in \VP{s_0}$ or $s'_0$ is obviously inconsistent iff $c^\ast \in \VP{s_0}^\ast$ or $s'_0$ is obviously inconsistent iff (by Lemma~\ref{lem:setup-involution}) $c^\ast \in \VP{s_0^\ast}$ is obviously inconsistent iff (by Lemma~\ref{lem:setup-disjoint-vp}) $\cdot,s_0^\ast \cup s'_0 \modelss c^\ast$.

\item The induction steps for $\neg (\alpha \lor \beta)$, $\ex x \alpha$, $\neg \ex x \alpha$, and $\neg \neg \alpha$ are trivial.
    \qedhere
\end{itemize}
\end{pf}

\begin{cor} \label{cor:objective-involution}
Let $\psi$ be objective and $\ast$ be a name involution.
Then $\entailss \psi$ iff $\entailss \psi^\ast$.
\end{cor}

\begin{pf}
Follows from Lemma~\ref{lem:objective-involution} for $s'_0 = \emptyset$.
\end{pf}

\begin{lem} \label{lem:lmod-objective-involution}
Let $\psi$ be objective, $\ast$ be a name involution, and $\Terms(s'_0)$ be disjoint from $\Terms(s_0) \cup \Terms(\gnd(\psi))$.
Then $s_0 \cup s'_0 \modelss \Lmod[k] \psi$ iff $s_0^\ast \cup s'_0 \modelss \Lmod[k] \psi^\ast$.
\end{lem}

\begin{pf}
We show $s_0 \cup s'_0,\cdot,v \cup v' \modelss \Lmod[k] \psi$ iff $s_0^\ast \cup s'_0,\cdot,v^\ast \cup v' \modelss \Lmod[k] \psi^\ast$ where $\Terms(v')$ are disjoint from $\Terms(s_0) \cup \Terms(\gnd(\psi)) \cup \Terms(v)$.
The proof is by induction on $k$.
\begin{itemize}
\item For the base case consider $k = 0$.
    By Lemma~\ref{lem:objective-involution}, $\cdot,s_0 \cup s'_0 \cup v \cup v' \modelss \psi$ iff $\cdot,s_0^\ast \cup s'_0 \cup v^\ast \cup v' \modelss \psi^\ast$.
    Moreover, $s_0 \cup s'_0 \cup v \cup v'$ is obviously/potentially inconsistent iff (by Lemma~\ref{lem:setup-disjoint-vp}) $s_0 \cup v$ or $s'_0 \cup v'$ is obviously/potentially inconsistent iff (by Lemma~\ref{lem:setup-involution}) $s_0^\ast \cup v^\ast$ or $s'_0 \cup v'$ is obviously/potentially inconsistent.
    Thus $s_0 \cup s'_0,\cdot,v \cup v' \modelss \Lmod[0] \psi$ iff $s_0^\ast \cup s'_0,\cdot,v^\ast \cup v' \modelss \Lmod[0] \psi^\ast$.

%
%
%

\item Now consider $k + 1$ and suppose $s_0 \cup s'_0,\cdot,v \cup v' \modelss \K[k] \psi$ iff $s_0^\ast \cup s'_0,\cdot,v^\ast \cup v' \modelss \K[k] \psi^\ast$.
    Then $s_0 \cup s'_0,\cdot,v \cup v' \modelss \K[k+1] \psi$ iff for some $t$ and all $n \in \Names_t$, $s_0 \cup s'_0,\cdot,v \cup v' \cup \{t \e n\} \modelss \K[k] \psi$ iff (by induction) for some $t$ and all $n \in \Names_t$, $s_0^\ast \cup s'_0,\cdot,(v \cup \{t \e n\})^\ast \cup v' \modelss \K[k] \psi^\ast$ or $s_0^\ast \cup s'_0,\cdot,v^\ast \cup (v' \cup \{t \e n\}) \modelss \K[k] \psi^\ast$ iff $s_0^\ast \cup s'_0,\cdot,v^\ast \cup v' \modelss \K[k+1] \psi^\ast$.

\item Now consider $k + 1$ and suppose $s_0 \cup s'_0,\cdot,v \cup v' \modelss \M[k] \psi$ iff $s_0^\ast \cup s'_0,\cdot,v^\ast \cup v' \modelss \M[k] \psi^\ast$.
    Then $s_0 \cup s'_0,\cdot,v \cup v' \modelss \M[k+1] \psi$ iff for some $t$ and $n \in \Names_t$, $s_0 \cup s'_0,\cdot,v \cup \{t \e n\} \cup v' \modelss \M[k] \psi$ or $s_0 \cup s'_0,\cdot,(v \uplus_{s_0} (t \e n)) \cup v' \modelss \M[k] \psi$ or $s_0 \cup s'_0,\cdot,v \cup (v' \uplus_{s'_0} (t \e n)) \modelss \M[k] \psi$ iff (by induction and Lemma~\ref{lem:setup-involution}) for some $t$ and all $n \in \Names_t$, $s_0^\ast \cup s'_0,\cdot,(v \cup \{t \e n\})^\ast \cup v' \modelss \M[k] \psi^\ast$ or $s_0^\ast \cup s'_0,\cdot,v \cup (v' \cup \{t \e n\}) \modelss \M[k] \psi^\ast$ or $s_0^\ast \cup s'_0,\cdot,(v^\ast \uplus_{s_0^\ast} (t^\ast \e n^\ast)) \cup v' \modelss \M[k] \psi$ or $s_0^\ast \cup s'_0,\cdot,v \cup (v' \uplus_{s'_0} (t \e n)) \modelss \M[k] \psi$ iff $s_0^\ast \cup s'_0,\cdot,v^\ast \cup v' \modelss \M[k+1] \psi^\ast$.
    \qedhere
\end{itemize}
\end{pf}

\begin{lem} \label{lem:lmod-objective-involution-kb-unchanged}
Let $\phi$ be proper\plus, $T = \Terms$ or $T = \Terms(\gnd(\alpha)) \supseteq \Terms(\gnd(\psi))$ for some $\alpha$ be a set of primitive terms, $\psi$ be objective, and $\ast$ be a name involution that leaves the names in $\phi$ unchanged.
Then $\gnd(\phi)|_T \modelss \Lmod[k] \psi$ iff $\gnd(\phi)|_T \modelss \Lmod[k] \psi^\ast$.
\end{lem}

\begin{pf}
First consider $T = \Terms$.
Then $\gnd(\phi)|_T = \WP{\gnd(\phi)}$.
Since $\gnd(\phi) = \gnd(\phi)^\ast$ and by Lemma~\ref{lem:setup-involution}, we have $\WP{\gnd(\phi)} = \WP{\gnd(\phi)}^\ast$.
Thus $\gnd(\phi)|_T = \gnd(\phi)|_T^\ast$ and the lemma follows from Lemma~\ref{lem:lmod-objective-involution}.

Now consider $T = \Terms(\gnd(\alpha))$ for some $\alpha$ such that $\Terms(\gnd(\alpha)) \supseteq \Terms(\gnd(\psi))$.
Let $s_0$ be the least set such that if $c \in \WP{\gnd(\phi)}$ and $c$ mentions a term from $\Terms(\gnd(\psi))$ or shares a term with another clause in $s_0$, then $c \in s_0$.
Let $s'_0$ be $\WP{\gnd(\phi)|_T} \setminus s_0$.
By construction, $\Terms(s'_0)$ is disjoint from $\Terms(s_0) \cup \Terms(\gnd(\psi))$ (*).
Furthermore, $s_0^\ast$ is the least set such that if $c \in \WP{\gnd(\phi)}^\ast$ and $c$ mentions a term from $\Terms(\gnd(\psi))^\ast$ or shares a term with another clause in $s_0^\ast$, then $c \in s_0^\ast$; by Lemma~\ref{lem:setup-involution} and since $\gnd(\phi) = \gnd(\phi)^\ast$, this is the least set such that if $c \in \WP{\gnd(\phi)}$ and $c$ mentions a term from $\Terms(\gnd(\psi))$ or shares a term with another clause in $s_0^\ast$, then $c \in s_0^\ast$; and hence $s_0 = s_0^\ast$ (**).
Then $\gnd(\phi)_T \modelss \Lmod[k] \psi$ iff $s_0 \cup s'_0 \modelss \Lmod[k] \psi$ iff (by (*) and Lemma~\ref{lem:lmod-objective-involution}) $s_0^\ast \cup s'_0 \modelss \Lmod[k] \psi^\ast$ iff (by (**)) $s_0 \cup s'_0 \modelss \Lmod[k] \psi^\ast$ iff $\gnd(\phi)|_T \modelss \Lmod[k] \psi^\ast$.
\end{pf}

\begin{lem} \label{lem:res}
Let $\phi$ be proper\plus, $T = \Terms$ or $T = \Terms(\gnd(\alpha)) \supseteq \Terms(\gnd(\psi))$ for some $\alpha$ be a set of primitive terms, and $\psi$ be objective.
Then $\gnd(\phi)|_T \modelss \Lmod[k] \psi^{x_1 \ldots x_j}_{n_1 \ldots n_j}$ iff $\entailss \RES{\phi,T}{\Lmod[k] \psi}^{x_1 \ldots x_j}_{n_1 \ldots n_j}$.
\end{lem}

\begin{pf}
By induction on $j$.
\begin{itemize}
\item For the base case, $\entailss \RES{\phi,T}{\Lmod[k] \psi}$ iff $\entailss \TRUE$ if $\gnd(\phi)|_T \modelss \Lmod[k] \psi$ and otherwise $\entailss \neg \TRUE$ iff $\gnd(\phi)|_T \modelss \Lmod[k] \psi$.
\item For the base case, suppose $j > 0$ and $\psi^{x_1}_{n_1}$ satisfies the lemma for every $n_1 \in \Names_{x_1}$.
    Consider $\RES{\phi,T}{\Lmod[k] \psi}^{x_1 \ldots x_j}_{n_1 \ldots n_j}$.

    If $n_1$ occurs in $\phi$ or $\psi$, then all but one of the disjuncts in $\RES{\phi,T}{\Lmod[k] \psi}^{x_1 \ldots x_j}_{n_1 \ldots n_j}$ are false, and we have $\entailss \RES{\phi,T}{\Lmod[k] \psi}^{x_1 \ldots x_j}_{n_1 \ldots n_j}$ iff $\entailss \RES{\phi,T}{\Lmod[k] \psi^x_{n_1}}^{x_2 \ldots x_j}_{n_2 \ldots n_j}$ iff (by induction) $\gnd(\phi)|_T \modelss \Lmod[k] \psi^{x_1 \ldots x_j}_{n_1 \ldots n_j}$.

    If $n_1$ does not occur in $\phi$ or $\psi$, then all but the last disjuncts in $\RES{\phi,T}{\Lmod[k] \psi}^{x_1 \ldots x_j}_{n_1 \ldots n_j}$ are false, and the last disjunct simplifies to $\entailss \RES{\phi,T}{\Lmod[k] \psi^{x_1}_{\hat{n}}}^{\hat{n}_{\phantom{1}} x_1 \ldots x_j}_{x_1 n_1 \ldots n_j}$.
    As $\RES{\phi,T}{\Lmod[k] \psi^{x_1}_{\hat{n}}}^{\hat{n}}_{x_1}$ mentions neither $\hat{n}$ nor $n_1$, we can apply an involution that swaps $\hat{n}$ and $n_1$ and leaves everything else unchanged, and $\entailss \RES{\phi,T}{\Lmod[k] \psi^{x_1}_{\hat{n}}}^{\hat{n}_{\phantom{1}} x_1 \ldots x_j}_{x_1 n_1 \ldots n_j}$ iff (by Corollary~\ref{cor:objective-involution}) $\entailss \RES{\phi,T}{\Lmod[k] \psi^{x_1}_{\hat{n}}}^{\hat{n}_{\phantom{1}} x_1 \ldots x_j}_{x_1 n_1^\ast \ldots n_j^\ast}$ iff (since $n_1^\ast \e \hat{n}$) iff $\entailss \RES{\phi,T}{\Lmod[k] \psi^{x_1}_{\hat{n}}}^{x_2 \ldots x_j}_{n_2^\ast \ldots n_j^\ast}$ iff (by induction) $\gnd(\phi)|_T \modelss \Lmod[k] \psi^{x_1 x_2 \ldots x_j}_{\hat{n}_{\phantom{1}} n_2^\ast \ldots n_j^\ast}$ iff (by Lemma~\ref{lem:lmod-objective-involution-kb-unchanged}) $\gnd(\phi)|_T \modelss \Lmod[k] \psi^{x_1 \ldots x_j}_{n_1 \ldots n_j}$.
    \qedhere
\end{itemize}
\end{pf}

\subsection{The $\RED{}{\cdot}$ Operator}

\begin{defi}
Let $\phi$ be proper\plus, $T$ be a set of primitive terms, and $\alpha$ without $\OO$.
Then $\RED{\phi,T}{\alpha}$ is defined as follows:
\begin{itemize}
\item $\RED{\phi,T}{t \e t'}$ is $t \e t'$;
\item $\RED{\phi,T}{\neg \alpha}$ is $\neg \RED{\phi,T}{\alpha}$;
\item $\RED{\phi,T}{(\alpha \lor \beta)}$ is $(\RED{\phi,T}{\alpha} \lor \RED{\phi,T}{\beta})$;
\item $\RED{\phi,T}{\ex x \alpha}$ is $\ex x \RED{\phi,T}{\alpha}$;
\item $\RED{\phi,T}{\Lmod[k] \alpha}$ is $\RES{\phi,T}{\Lmod[k] \RED{\phi,T}{\alpha}}$;
\item $\RED{\phi,T}{\G \alpha}$ is $\RED{\phi,T \cap \Terms(\gnd(\alpha))}{\alpha}$.
\end{itemize}
\end{defi}

\begin{lem} \label{lem:representation}
Let $\phi$ be proper\plus, $T = \Terms$ or $T = \Terms(\gnd(\alpha)) \supseteq \Terms(\gnd(\sigma))$ for some $\alpha$ be a set of primitive terms, and $\sigma$ be subjective and without $\OO$.
Then $\gnd(\phi)|_T \modelss \sigma^{x_1 \ldots x_j}_{n_1 \ldots n_j}$ iff $\entailss \RED{\phi,T}{\sigma}{}^{x_1 \ldots x_j}_{n_1 \ldots n_j}$.
\end{lem}

\begin{pf}
We show $\gnd(\phi)|_T,s \modelss \alpha^{x_1 \ldots x_j}_{n_1 \ldots n_j}$ iff $\cdot,s \entailss \RED{\phi,T}{\alpha}{}^{x_1 \ldots x_j}_{n_1 \ldots n_j}$ by induction on $|\alpha|$.
\begin{itemize}
\item The non-modal cases are trivial since $\RED{}{\cdot}$ does not introduce any new clauses.

\item Consider $\Lmod[k] \alpha$.
    Then $\gnd(\phi)|_T \modelss \Lmod[k] \alpha^{x_1 \ldots x_j}_{n_1 \ldots n_j}$ iff (by reducing $\Lmod[k]$) for certain sets of split literals $v$, $\gnd(\phi)|_T \cup v$ is obviously/not potentially inconsistent or/and $\cdot,\gnd(\phi)|_T \cup v \modelss \alpha^{x_1 \ldots x_j}_{n_1 \ldots n_j}$ iff (by induction) for these $v$, $\gnd(\phi)|_T \cup v$ is obviously/not potentially inconsistent or/and $\cdot,\gnd(\phi)|_T \cup v \modelss \RED{\phi,T}{\alpha}{}^{x_1 \ldots x_j}_{n_1 \ldots n_j}$ iff (by reversing the steps to reduce $\Lmod[k]$) $\gnd(\phi)|_T \modelss \Lmod[k] \RED{\phi,T}{\alpha}{}^{x_1 \ldots x_j}_{n_1 \ldots n_j}$ iff (since $\RED{\phi,T}{\alpha}{}^{x_1 \ldots x_j}_{n_1 \ldots n_j}$ is objective, by Lemma~\ref{lem:res}) $\gnd(\phi)|_T \modelss \RES{\phi,T}{\Lmod[k] \RED{\phi,T}{\alpha}}^{x_1 \ldots x_j}_{n_1 \ldots n_j}$ iff $\gnd(\phi)|_T \modelss \RED{\phi,T}{\Lmod[k] \alpha}{}^{x_1 \ldots x_j}_{n_1 \ldots n_j}$.

\item Consider $\G \alpha$.
    Then $\gnd(\phi)|_T \modelss \G \alpha^{x_1 \ldots x_j}_{n_1 \ldots n_j}$ iff $(\gnd(\phi)|_T)_{\Terms(\gnd(\alpha))} \modelss \alpha^{x_1 \ldots x_j}_{n_1 \ldots n_j}$ iff $\gnd(\phi)|_{T \cap \Terms(\gnd(\alpha))} \modelss \alpha^{x_1 \ldots x_j}_{n_1 \ldots n_j}$ iff (by induction) $\entailss \RED{\phi,T \cap \Terms(\gnd(\alpha))}{\alpha}{}^{x_1 \ldots x_j}_{n_1 \ldots n_j}$ iff $\entails \RED{\phi,T}{\G \alpha}{}^{x_1 \ldots x_j}_{n_1 \ldots n_j}$.
    \qedhere
\end{itemize}
\end{pf}

\begin{repthm}[Representation theorem]{thm:representation}
Let $\phi$ be proper\plus, $\sigma$ be subjective and without $\OO$.
Then $\OO \phi \entailss \sigma$ iff $\entailss \RED{\phi,\Terms}{\sigma}$.
\end{repthm}

\begin{pf}
$\OO \phi \entailss \sigma$ iff (by Theorem~\ref{thm:o}) $\gnd(\phi) \modelss \sigma$ iff (by Lemma~\ref{lem:setup-fix-modelss}) $\gnd(\phi)|_\Terms \modelss \sigma$ iff (by Lemma~\ref{lem:representation}) $\entailss \RED{\phi,\Terms}{\sigma}$.
\end{pf}

\section{Decidability} \label{app:decidable}

This section proves the decidability result Theorem~\ref{thm:decidable}.
First we show that only finitely many literals need to be considered for both $\K[k]$ and $\M[k]$.
Next we show that finite groundings of proper\plus\ knowledge bases suffice.
Then we put these results together to define a decision procedure.

\begin{defi}
For a term $t$, let $\Vars_t(\alpha)$ be the maximum number of free variables in any subformula of $\alpha$.
Then we define the following sets of names:
\begin{itemize}
\item $\Names_{t\bullet}(\alpha) = \Names_t(\alpha) \cap \{n \mid \text{$t \e n$ or $t \n n$ is a subformula of an element of $\gnd(\alpha)$}\}$;
\item $\NamesP{p}_t(\alpha) = \Names_t(\alpha) \cup \{n_1,\ldots,n_p \in \Names_t \setminus \Names_t(\alpha)\}$;
\item $\NamesP{p}_{t\bullet}(\alpha) = \Names_{t\bullet}(\alpha) \cup \{n_1,\ldots,n_p \in \Names_t \setminus \Names_t(\alpha)\}$;
\item $\NamesPV{p}_t(\alpha) = \NamesP{\Vars_t(\alpha)+p}_t(\alpha)$;
\item $\NamesPV{p}_{t\bullet}(\alpha) = \NamesP{\Vars_t(\alpha)+p}_{t\bullet}(\alpha)$.
\end{itemize}
(To make $\NamesP{p}_t(\alpha), \NamesP{p}_{t\bullet}(\alpha)$ well-defined, we assume $n_p$ are the minimal names w.r.t.\ some preorder $<$ that do not occur in $\Names_t(\alpha)$).
Let $\NamesP{p}(\alpha) = \bigcup_t \NamesP{p}_t(\alpha)$ and $\NamesPV{p}(\alpha) = \bigcup_t \NamesPV{p}(\alpha)$.
These definitions naturally extend to sets of formulas and terms; when the set is finite, we sometimes omit the brackets.
\end{defi}

\subsection{Finite Splitting for $\K[k]$}

\begin{lem} \label{lem:term-restricted-name}
Let $\phi$ be proper\plus, $\psi$ be objective, and $n_1 \in \Names_t \setminus \Names_{t\bullet}(\phi,\psi,t)$ and $n_2 \in \Names_t \setminus \Names_t(\phi,\psi,t)$.
Then $\gnd(\phi),\cdot,\{t \e n_1\} \modelss \Lmod[k] \psi$ iff $\gnd(\phi),\cdot,\{t \e n_2\} \modelss \Lmod[k] \psi$.
\end{lem}

\begin{pf}
Since $n_1 \in \Names_t \setminus \NamesP0_{t\bullet}(\phi,\psi,t)$, we have either $n_1 \in \Names_t \setminus \Names_t(\phi,\psi,t)$ or $n_1 \in \Names_t(\phi,\psi,t)$ but $t \e n_1$ and $t \n n_1$ are no subformulas of $\gnd(\phi,\psi,t)$.

First suppose $n_1 \in \Names_t \setminus \Names_t(\phi,\psi,t)$.
Let $\ast$ be the name involution that swaps $n_1$ and $n_2$ and leaves all other names unchanged.
Then $\gnd(\phi),\cdot,\{t \e n_1\} \modelss \Lmod[k] \psi$ iff $\gnd(\phi)^\ast,\cdot,\{t \e n_1\}^\ast \modelss \Lmod[k] \psi^\ast$ iff $\gnd(\phi),\cdot,\{t \e n_2\} \modelss \Lmod[k] \psi$.

Now suppose $n_1 \in \Names_t(\phi,\psi,t)$ but $t \e n_1$ and $t \n n_1$ are no subformulas of $\gnd(\phi,\psi,t)$.
First we show that if $\UP{\gnd(\phi) \cup \{t \e n_1\}} \setminus \{t \e n_1\} = \UP{\gnd(\phi) \cup \{t \e n_2\}} \setminus \{t \e n_2\}$, and $c \in (\UP{\gnd(\phi) \cup \{t \e n_1\}} \setminus \{t \e n_1\}) \cup (\UP{\gnd(\phi) \cup \{t \e n_2\}} \setminus \{t \e n_1\})$ contains neither $t \e n_1$, $t \n n_1$, $t \e n_2$, nor $t \n n_2$ by induction on the length of the derivation of $c$.
\begin{itemize}
\item For the base case, $c \in \UP[0]{\gnd(\phi) \cup \{t \e n_1\}} \setminus \{t \e n_1\}$ iff $c$ is a valid literal or $\gnd(\phi)$ iff $c \in \UP[0]{\gnd(\phi) \cup \{t \e n_2\}} \setminus \{t \e n_2\}$.
    Moreover, $c$ contains neither $t \e n_1$ nor $t \n n_2$.
    Finally suppose $c$ contains $t \e n_2$ or $t \n n_2$; since neither is valid, $c \in \gnd(\phi)$, and since $n_2 \in \Terms_t \setminus \Terms_t(\phi,\psi,t)$, there $c$ must contain a literal $t \e x$ or $t \n x$; then, however, $c$ also mentions $t \e n_1$ or $t \n n_1$, which contradicts the assumption.
    Hence $c$ also mentions neither $t \e n_2$ nor $t \n n_2$.

\item Now suppose $c \in \UP[k+1]{\gnd(\phi) \cup \{t \e n_1\}} \setminus (\UP[k]{\gnd(\phi) \cup \{t \e n_1\}}$.
    Then $c$ is the resolution of $c', \ell \in \UP[k]{\gnd(\phi) \cup t \e n_1}$.
    By induction, we have four cases concerning $c', \ell$:
    \begin{itemize}
    \item Case 1: $c', \ell \in \UP[k]{\gnd(\phi) \cup \{t \e n_2\}}$ and $c', \ell$ contain neither $t \e n_1$, $t \n n_1$, $t \e n_2$, nor $t \n n_2$.
        Then also $c \in \UP[k+1]{\gnd(\phi) \cup \{t \e n_2\}}$, and $c$ contains neither $t \e n_1$, $t \n n_1$, $t \e n_2$, nor $t \n n_2$.
    \item Case 2: $c' \in \UP[k]{\gnd(\phi) \cup \{t \e n_2\}}$ and $c'$ contains neither $t \e n_1$, $t \n n_1$, $t \e n_2$, nor $t \n n_2$, and $\ell$ is $t \e n_1$.
        Then all literals in $c'$ complementary to $\ell$ are of the form $t \e n$ for $n$ distinct from $n_1$.
        By assumption, all these literals are distinct from $t \e n_2$.
        Hence $c$ is the unit propagation of $c'$ and $t \e n_2$, and hence $c \in \UP[k+1]{\gnd(\phi) \cup \{t \e n_2\}}$ and $c$ contains neither $t \e n_1$, $t \n n_1$, $t \e n_2$, nor $t \n n_2$.
    \item Case 3: $\ell \in \UP[k]{\gnd(\phi) \cup \{t \e n_2\}}$ and $\ell$ is neither $t \e n_1$, $t \n n_1$, $t \e n_2$, nor $t \n n_2$, and $c'$ is $t \e n_1$.
        This is just a special case of Case 2.
    \item Case 4: $c', \ell$ both are $t \e n_1$.
        This contradicts the assumption that $c$ is the unit propagation of $c', \ell$.
    \end{itemize}

\item Now suppose $c \in \UP[k+1]{\gnd(\phi) \cup \{t \e n_2\}} \setminus (\UP[k]{\gnd(\phi) \cup \{t \e n_2\}}$.
    Then $c$ is the resolution of $c', \ell \in \UP[k]{\gnd(\phi) \cup t \e n_2}$.
    By induction, we have two different cases concerning $c', \ell$:
    \begin{itemize}
    \item Case 1: $c', \ell \in \UP[k]{\gnd(\phi) \cup \{t \e n_1\}}$ and $c', \ell$ contain neither $t \e n_1$, $t \n n_1$, $t \e n_2$, nor $t \n n_2$.
        Then also $c \in \UP[k+1]{\gnd(\phi) \cup \{t \e n_1\}}$, and $c$ contains neither $t \e n_1$, $t \n n_1$, $t \e n_2$, nor $t \n n_2$.
    \item Case 2: $c' \in \UP[k]{\gnd(\phi) \cup \{t \e n_1\}}$ and $c'$ contains neither $t \e n_1$, $t \n n_1$, $t \e n_2$, nor $t \n n_2$, and $\ell$ is $t \e n_2$.
        Then all literals in $c'$ complementary to $\ell$ are of the form $t \e n$ for $n$ distinct from $n_2$.
        By assumption, all these literals are distinct from $t \e n_1$.
        Hence $c$ is the unit propagation of $c'$ and $t \e n_1$, and hence $c \in \UP[k+1]{\gnd(\phi) \cup \{t \e n_1\}}$ and $c$ contains neither $t \e n_1$, $t \n n_1$, $t \e n_2$, nor $t \n n_2$.
    \item Case 3: $\ell \in \UP[k]{\gnd(\phi) \cup \{t \e n_1\}}$ and $\ell$ is neither $t \e n_1$, $t \n n_1$, $t \e n_2$, nor $t \n n_2$, and $c'$ is $t \e n_2$.
        This is just a special case of Case 2.
    \item Case 4: $c', \ell$ both are $t \e n_2$.
        This contradicts the assumption that $c$ is the unit propagation of $c', \ell$.
    \end{itemize}
\end{itemize}
Since no clause in $\UP{\gnd(\phi) \cup \{t \e n_i\}} \setminus \{t \e n_i\}$ mentions $t \n n_i$, both $\gnd(\phi) \cup \{t \e n_1\}$ is potentially inconsistent iff $\gnd(\phi) \cup \{t \e n_2\}$ is not potentially inconsistent.
An induction on $k$ and subinduction on $|\psi|$ then shows that $\gnd(\phi),\cdot,\{t \e n_1\} \modelss \Lmod[k] \psi$ iff $\gnd(\phi),\cdot,\{t \e n_2\} \modelss \Lmod[k] \psi$.
\end{pf}

\begin{lem} \label{lem:split-objective-kb-k}
Let $\phi$ be proper\plus\ and $\psi$ be objective.
Then $\gnd(\phi) \modelss \K[k+1] \psi$ iff for some $t \in \Terms$, for all $n \in \Names_t$, $\gnd(\phi \land t \e n) \modelss \K[k] \psi$.
\end{lem}

\begin{pf}
We show by induction on $k$ that $\gnd(\phi),\cdot,\{t \e n\} \cup v \modelss \K[k] \psi$ iff $\gnd(\phi \land t \e n),\cdot,v \modelss \K[k] \psi$.
\begin{itemize}
\item The base case follows by a simple subinduction on $|\psi|$ since $\gnd(\phi) \cup \{t \e n\} \cup v = \gnd(\phi \land t \e n) \cup v$.
\item For the induction step, $\gnd(\phi),\cdot,\{t \e n\} \cup v \modelss \K[k+1] \psi$ iff for some $t' \in \Terms$, for all $n' \in \Names_{t'}$, $\gnd(\phi),\cdot,\{t \e n\} \cup v \cup \{t' \e n'\} \modelss \K[k] \psi$ iff (by induction) for some $t' \in \Terms$, for all $n' \in \Names_{t'}$, $\gnd(\phi \land t \e n),\cdot,v \cup \{t' \e n'\} \modelss \K[k] \psi$ iff $\gnd(\phi \land t \e n),\cdot,v \modelss \K[k+1] \psi$.
\end{itemize}
Since $\gnd(\phi) \modelss \K[k+1] \psi$ iff for some $t \in \Terms$, for all $n \in \Names_t$, $\gnd(\phi),\cdot,\emptyset \cup \{t \e n\} \modelss \K[k] \psi$, the lemma follows by the above statement.
\end{pf}

\begin{lem} \label{lem:split-objective-k}
Let $\phi$ be proper\plus, $\psi$ be objective.
Then $\gnd(\phi) \modelss \K[k+1] \psi$ iff for some $t \in \Terms(\gnd_{\NamesPV0(\phi,\psi)}(\phi,\psi))$, for all $n \in \NamesP1_{t\bullet}(\phi,\psi,t)$, $\gnd(\phi \land t \e n) \modelss \K[k] \psi$.
\end{lem}

\begin{pf}
First we show that $\gnd(\phi) \modelss \K[k+1] \psi$ iff for some $t \in \Terms(\gnd_{\NamesPV0(\phi,\psi)}(\phi, \psi))$, for all $n \in \Names_t$, $\gnd(\phi \land t \e n) \modelss \K[k] \psi$.
The if direction immediately follows by the semantics.
Conversely, suppose $\gnd(\phi) \modelss \K[k+1] \psi$.
Then for some $t \in \Terms$, for all $n \in \Names_t$, $\gnd(\phi),\cdot,\{t \e n\} \modelss \K[k] \psi$.
Then there is such a $t$ with $t \in \Terms(\gnd(\phi,\psi))$, for otherwise $t \e n$ cannot be complementary to or subsume any literal occurring in $\gnd(\phi,\psi)$.
In particular, then $t$ mentions at most $\Vars_{n'}(\phi,\psi)$ names of the same sort as $n'$ that do not occur in $\phi$ or $\psi$.
Let $\ast$ be a name involution that swaps these $n' \in \Names(t) \setminus \Names(\phi,\psi)$ with some $n'{}^\ast \in \NamesPV0_{n'}(\phi,\psi)$ and leaves all other names unchanged.
Then $t^\ast \in \Terms(\gnd_{\NamesPV0(\phi,\psi)}(\phi,\psi))$, and for every $n \in \Names_t$, $\gnd(\phi),\cdot,\{t \e n\} \modelss \K[k] \psi$ iff for every $n \in \Names_t$, $\gnd(\phi)^\ast,\cdot,\{t^\ast \e n^\ast\} \modelss \K[k] \psi$ iff for every $n \in \Names_t$, $\gnd(\phi),\cdot,\{t^\ast \e n\} \modelss \K[k] \psi$ iff (by Lemma~\ref{lem:split-objective-kb-k}) for every $n \in \Names_t$, $\gnd(\phi \land t^\ast \e n) \modelss \K[k] \psi$.

Next we show that $\gnd(\phi) \modelss \K[k+1] \psi$ iff for some $t \in \Terms$, for all $n \in \NamesP1_t(\phi,\psi,t)$, $\gnd(\phi \land t \e n) \modelss \K[k] \psi$.
The only-if direction immediately follows by the semantics.
Conversely, suppose for some $t \in \Terms$, for all $n \in \NamesP1_{t\bullet}(\phi,\psi,t)$, $\gnd(\phi \land t \e n) \modelss \K[k] \psi$.
By Lemma~\ref{lem:split-objective-kb-k}, for all $n \in \NamesP1_{t\bullet}(\phi,\psi,t)$, $\gnd(\phi),\cdot,\{t \e n\} \modelss \K[k] \psi$.
Let $\{\hat{n}\} = \NamesP1_{t\bullet}(\phi,\psi,t) \setminus \Names_{t\bullet}(\phi,\psi,t)$; then $\hat{n} \in \Names_t \setminus \Names_t(\phi,\psi,t)$ and $\gnd(\phi),\cdot,\{t \e \hat{n}\} \modelss \K[k] \psi$, and by Lemma~\ref{lem:term-restricted-name}, for all $n \in \Names_t \setminus \Names_{t\bullet}(\phi,\psi,t)$, $\gnd(\phi),\cdot,\{t \e n\} \modelss \K[k] \psi$.
Hence for every name $n \in \Names_t$, $\gnd(\phi),\cdot,\{t \e n\} \modelss \K[k] \psi$.
Thus $\gnd(\phi) \modelss \K[k+1] \psi$.
\end{pf}

\subsection{Finite Splitting for $\M[k]$}

\begin{defi}
Let $\phi$ be proper\plus\ and $\ell$ be a literal $t \e n$.
Let $\Phi$ be the least set such that if $\ell, \hat\ell$ are isomorphic and $\neg \hat\ell \notin \UP{\gnd(\phi)}$, then $\bigvee_{1 \leq i \leq j, n \in \Names(\phi)} x_i \e n \lor \bigvee_{1 \leq i_1 < i_2 \leq j} x_{i_1} \e x_{i_2} \lor \hat\ell\,^{n_1 \ldots n_j}_{x_1 \ldots x_j} \in \Phi$ where $\{n_1,\ldots,n_j\} \in \Names(\hat\ell) \setminus \Names(\phi)$ and $x_1,\ldots,x_j$ are variables of corresponding sort.
Then $\phi \uplus \ell$ is defined as $\{\phi\} \cup \Phi$.
\end{defi}

\begin{lem} \label{lem:add-lits-well-defined}
Let $\phi$ be proper\plus\ and $\ell$ be a literal $t \e n$.
Then $\phi \uplus \ell$ is finite (modulo variable renamings) and a well-defined formula, $\Names(\phi \uplus \ell) = \Names(\phi)$, and $\Vars(\phi \uplus \ell) \leq \max \{\Vars(\phi), |t|+1\}$, where $|t|+1$ is the arity of $t$.
\end{lem}

\begin{pf}
Assume the notation from the definition of $\phi \uplus \ell$.
Since $\Names(\phi)$ is finite, every element in $\Phi$ is a well-defined formula.
Since there are only finitely many variables in $\ell^{n_1 \ldots n_j}_{x_1 \ldots x_j}$, the set $\Phi$ must be finite as well (modulo variable renamings).
Hence $\phi \uplus \ell$ is a well-defined formula.
Since $\Phi$ introduces no new names, $\Names(\phi \uplus \ell) = \Names(\phi)$, and since the clauses in $\Phi$ contain at most $|t|+1$ variables, $\Vars(\phi \uplus \ell) \leq \max\{\Vars(\phi), |t|+1\}$.
\end{pf}

\begin{lem} \label{lem:setup-add-lits-new}
Let $\phi$ be proper\plus\ and $\ell$ be a literal $t \e n$.
Then $\WP{\gnd((\phi \uplus \ell) \setminus \{\phi\})} = \emptyset \uplus_{\gnd(\phi)} \ell$.
\end{lem}

\begin{pf}
First we show the $\subseteq$ direction.
By construction, every clause in $\UP{\gnd((\phi \uplus \ell) \setminus \{\phi\})}$ is either a unit clause or subsumed by a unit clause.
Hence $\WP{\gnd((\phi \uplus \ell) \setminus \{\phi\})}$ only contains literals $\tilde\ell$ of the form $(\hat\ell\,^{n_1 \ldots n_j}_{x_1 \ldots x_j})^{x_1 \ldots x_j}_{n'_1 \ldots n'_j}$, where $\hat\ell$ is isomorphic to $\ell$, $\neg \hat\ell \notin \UP{\gnd(\phi)}$, $\{n_1,\ldots,n_j\} = \Names(\hat\ell) \setminus \Names(\phi)$, $n'_i \notin \Names(\phi)$, and $n'_{i_1} \neq n'_{i_2}$ for $i_1 \neq i_2$.
Let $\ast$ be the involution that swaps $n'_i$ and $n_i$.
Then $\hat\ell$ is identical to $\tilde\ell^\ast$, so $\hat\ell, \tilde\ell$ are isomorphic.
Moreover, $\neg \hat\ell \notin \UP{\gnd(\phi)}$, so $\neg \tilde\ell \notin \UP{\gnd(\phi)}^\ast$, and by Lemma~\ref{lem:setup-involution}, $\neg \tilde\ell \notin \UP{\gnd(\phi)}$.
Hence, $\tilde\ell \in \emptyset \uplus_{\gnd(\phi)} \ell$.

For the $\supseteq$ direction suppose $\hat\ell \in \emptyset \uplus_{\gnd(\phi)} \ell$.
Then $\ell, \hat\ell$ are isomorphic and $\neg \hat\ell \notin \VP{\gnd(\phi)}$.
Then $\bigvee_{1 \leq i \leq j, n \in \Names(\phi)} x_i \e n \lor \bigvee_{1 \leq i_1 < i_2 \leq j} x_{i_1} \e x_{i_2} \lor \hat\ell\,^{n_1 \ldots n_j}_{x_1 \ldots x_j} \in (\phi \uplus \ell) \setminus \{\phi\}$ where $\{n_1,\ldots,n_j\} \in \Names(\hat\ell) \setminus \Names(\phi)$ and $x_1,\ldots,x_j$ are variables of corresponding sort, and hence $\hat\ell \in \UP{\gnd((\phi \uplus \ell) \setminus \{\phi\})}$.
\end{pf}

\begin{lem} \label{lem:setup-add-lits}
Let $\phi$ be proper\plus.
Then $\VP{\gnd(\phi \uplus (t \e n))} = \VP{\gnd(\phi) \cup (\emptyset \uplus_{\gnd(\phi)} (t \e n))}$.
\end{lem}

\begin{pf}
By Lemma~\ref{lem:setup-add-lits-new}, $\VP{\gnd(\phi) \cup (\emptyset \uplus_{\gnd(\phi)} \ell)} = \VP{\gnd(\phi) \cup \WP{\gnd((\phi \uplus \ell) \setminus \{\phi\})}}$, and by Lemma~\ref{lem:setup-vp-wp-invariant}, $\VP{\gnd(\phi) \cup \WP{\gnd((\phi \uplus \ell) \setminus \{\phi\})}} = \VP{\gnd(\phi) \cup \gnd((\phi \uplus \ell) \setminus \{\phi\})} = \VP{\gnd(\phi \uplus \ell)}$.
\end{pf}

\begin{lem} \label{lem:split-objective-kb-m}
Let $\phi$ be proper\plus\ and $\psi$ be objective.
Then $\gnd(\phi) \modelss \M[k+1] \psi$ iff for some $t \in \Terms$ and $n \in \Names_t$, $\gnd(\phi \land t \e n) \modelss \M[k] \psi$ or $\gnd(\phi \uplus (t \e n)) \modelss \M[k] \psi$.
\end{lem}

\begin{pf}
Precisely analogous to the proof of Lemma~\ref{lem:split-objective-kb-k}, we $\gnd(\phi),\cdot,\{t \e n\} \cup v \modelss \M[k] \psi$ iff $\gnd(\phi \land t \e n),\cdot,v \modelss \M[k] \psi$.
Additionally, we show by induction on $k$ that $\gnd(\phi),\cdot,(\emptyset \uplus_{\gnd(\phi)} (t \e n)) \cup v \modelss \M[k] \psi$ iff $\gnd(\phi \uplus (t \e n)),\cdot,v \modelss \M[k] \psi$.
\begin{itemize}
\item The base case follows by a simple subinduction on $|\psi|$ since $\VP{\gnd(\phi) \cup (\emptyset \uplus_{\gnd(\phi)} (t \e n)) \cup v} = \VP{\gnd(\phi \uplus (t \e n)) \cup v}$ by Lemmas \ref{lem:setup-vp-wp-invariant} and \ref{lem:setup-add-lits}.
\item For the induction step, $\gnd(\phi),\cdot,(\emptyset \uplus_{\gnd(\phi)} (t \e n)) \cup v \modelss \M[k+1] \psi$ iff for some $t' \in \Terms$ and $n' \in \Names_{t'}$, $\gnd(\phi),\cdot,(\emptyset \uplus_{\gnd(\phi)} (t \e n)) \cup v \cup \{t' \e n'\} \modelss \M[k] \psi$ or $\gnd(\phi),\cdot,((\emptyset \uplus_{\gnd(\phi)} (t \e n)) \cup v) \uplus_{\gnd(\phi)} (t' \e n') \modelss \M[k] \psi$ iff (by induction and definition of $\uplus_{\gnd(\phi)}$) for some $t' \in \Terms$ and $n' \in \Names_{t'}$, $\gnd(\phi \uplus (t \e n)),\cdot,v \cup \{t' \e n'\} \modelss \M[k] \psi$ or $\gnd(\phi),\cdot,v \uplus_{\gnd(\phi)} (t' \e n') \modelss \M[k] \psi$ iff $\gnd(\phi \uplus (t \e n),\cdot,v \modelss \M[k+1] \psi$.
\end{itemize}
Since $\gnd(\phi) \modelss \M[k+1] \psi$ iff for some $t \in \Terms$ and $n \in \Names_t$, $\gnd(\phi),\cdot,\emptyset \cup \{t \e n\} \modelss \M[k] \psi$ or $\gnd(\phi),\cdot,\emptyset \uplus_{\gnd(\phi)} (t \e n) \modelss \M[k] \psi$, the lemma follows by the above statement.
\end{pf}

\begin{lem} \label{lem:split-objective-m}
Let $\phi$ be proper\plus, $\psi$ be objective.
Then $\gnd(\phi) \modelss \M[k+1] \psi$ iff for some $t \in \Terms(\gnd_{\NamesPV0(\phi,\psi)}(\phi,\psi))$ and some $n \in \NamesP1_{t\bullet}(\phi,\psi,t)$, $\gnd(\phi \land t \e n) \modelss \M[k] \psi$ or $\gnd(\phi \uplus (t \e n)) \modelss \M[k] \psi$.
\end{lem}

\begin{pf}
The if direction immediately follows from the definition of the semantics.
For the converse direction, suppose $\gnd(\phi) \modelss \M[k+1] \psi$.
Then for some $t \in \Terms$ and $n \in \Names_t$, $\gnd(\phi),\cdot,\{t \e n\} \modelss \M[k] \psi$ or $\gnd(\phi),\cdot,\emptyset \uplus_{\gnd(\phi)} (t \e n) \modelss \M[k] \psi$.
Then there is such a $t$ with $t \in \Terms(\gnd(\phi,\psi))$ and $n \in \Names(\gnd(\phi,\psi))$, for otherwise $t \e n$ cannot be complementary to or subsume any literal occurring in $\gnd(\phi,\psi)$.
In particular, then $t$ mentions at most $\Vars_{n'}(\phi,\psi)$ names of the same sort as $n'$ that do not occur in $\phi$ or $\psi$.
Let $\ast$ be a name involution that swaps these $n' \in \Names(t) \setminus \Names(\phi,\psi)$ with some $n'{}^\ast \in \NamesPV0_{n'}(\phi,\psi)$ and swaps $n$ in case it is not in $\Names(t)$ with the name $\NamesP1_{t^\ast\bullet}(\phi,\psi,t^\ast) \setminus \Names_{t^\ast\bullet}(\phi,\psi,t^\ast)$ and leaves all other names unchanged.
Then $t^\ast \in \Terms(\gnd_{\NamesPV0(\phi,\psi)}(\phi,\psi))$ and $n^\ast \in \NamesP1_{t^\ast\bullet}(\phi,\psi,t^\ast)$, and $\gnd(\phi),\cdot,\{t \e n\} \modelss \M[k] \psi$ or $\gnd(\phi),\cdot,\emptyset \uplus_{\gnd(\phi)} (t \e n) \modelss \M[k] \psi$ iff $\gnd(\phi)^\ast,\cdot,\{t^\ast \e n^\ast\} \modelss \M[k] \psi^\ast$ or $\gnd(\phi)^\ast,\cdot,\emptyset \uplus_{\gnd(\phi)^\ast} (t^\ast \e n^\ast) \modelss \M[k] \psi^\ast$ iff $\gnd(\phi),\cdot,\{t^\ast \e n^\ast\} \modelss \M[k] \psi$ or $\gnd(\phi),\cdot,\emptyset \uplus_{\gnd(\phi)} (t^\ast \e n^\ast) \modelss \M[k] \psi$ iff (by Lemma~\ref{lem:split-objective-kb-m}) $\gnd(\phi \land t^\ast \e n^\ast) \modelss \M[k] \psi$ or $\gnd(\phi \uplus t^\ast \e n^\ast) \modelss \M[k] \psi$.
\end{pf}

\subsection{Finite Setup}

\begin{lem} \label{lem:setup-gnd-finite}
Let $\phi$ be proper\plus, $\alpha$ be some formula, and $c$ be a clause.
Suppose $|\Names_t(c) \setminus \Names_t(\phi,\alpha)| \leq \Vars_t(\phi)$ for every term $t$.
Let $\ast$ be an involution that swaps every $n \in \Names(c) \setminus \Names(\phi,\alpha)$ with a name $n^\ast \in \NamesPV1_n(\phi,\alpha) \setminus \Names_n(\phi,\alpha)$ and leaves all names in $\Names(\phi,\alpha)$ unchanged.
Then $c \in \UP{\gnd(\phi)}$ implies $c^\ast \in \UP{\gnd_{\NamesPV1(\phi,\alpha)}(\phi)}$.
\end{lem}

\begin{pf}
We show by induction on $k$ that when $c \in \Min{\UP[k]{\gnd(\phi)}}$ and $\ast$ is a bijection that swaps every $n \in \Names(c) \setminus \Names(\phi,\alpha)$ with an $n^\ast \in \NamesPV1_n(\phi,\alpha) \setminus \Names_n(\phi,\alpha)$ and leaves the rest unchanged, then $c^\ast \in \UP[k]{\gnd_{\NamesPV1(\phi,\alpha)}(\phi)}$.
\begin{itemize}
\item First consider the base case $k = 0$.
    If $c$ is a valid literal, then $c^\ast$ is a valid literal as well, and hence $c^\ast \in \UP{\gnd_{\NamesPV1(\phi,\alpha)}(\phi)}$.
    Otherwise, $c \in \gnd(\phi)$, so $c = (c_i)^{x_1 \ldots x_j}_{n_1 \ldots n_j}$ for some $c_i$ and $n_1,\ldots,n_j$.
    Then $c^\ast = (c_i)^{x_1 \ldots x_j}_{n_1^\ast \ldots n_j^\ast}$, and hence $c^\ast \in \UP{\gnd_{\NamesPV1(\phi,\alpha)}(\phi)}$.

\item Now suppose the claim holds for $k$ and suppose $c \in \Min{\UP[k+1]{\gnd(\phi)}}$ and $\ast$ is a bijection that swaps every $n \in \Names(c) \setminus \Names(\phi,\alpha)$ with an $n^\ast \in \NamesPV1_n(\phi,\alpha) \setminus \Names_n(\phi,\alpha)$ and leaves the rest unchanged.
    Now suppose $c$ is the unit propagation of $c', \ell \in \UP{\gnd(\phi)}$.
    By Lemmas \ref{lem:literal-valid} and \ref{lem:literal-complement}, either $c'$ or $\ell$ must not be valid; without loss of generality suppose that $c'$ is not valid.
    We have the following cases:
    \begin{enumerate}
    \item $\ell$ is of the form $t \n n_1$ and $c'$ contains $t \e n_1$, or alternatively $\ell$ is of the form $t \e n_1$ and $c'$ contains $t \n n_1$.
        Let $\circ$ be the involution such that $n^\circ$ is $n^\ast$ for every $n \in \Names(c) \setminus \Names(\phi,\alpha)$ and that furthermore swaps every $n \in \Names(\ell) \setminus \Names(\phi,\alpha)$ with $n^\circ \in \NamesPV1_n(\phi,\alpha) \setminus \Names_n(\phi,\alpha)$ and leaves the rest unchanged.
        Such an involution exists because $\Names_t(c') \supseteq \Names_t(\ell)$ and $|\Names_t(c') \setminus \Names_t(\phi,\alpha)| \leq |\NamesPV1_t(\phi,\alpha) \setminus \Names_t(\phi,\alpha)|$.
        Then $\circ$ is an involution that swaps all $n \in \Names(c') \setminus \Names(\phi,\alpha)$ with $n^\circ \in \NamesPV1_n(\phi,\alpha) \setminus \Names_n(\phi,\alpha)$.
        By induction, $c'{}^\circ, \ell^\circ \in \UP[k]{\gnd_{\NamesPV1(\phi,\alpha)}(\phi)}$.
        Since $c^\ast$ is the unit propagation of $c'{}^\circ$ and $\ell^\circ$, we have $c^\ast \in \UP[k+1]{\gnd_{\NamesPV1(\phi,\alpha)}(\phi)}$.

    \item $\ell$ is of the form $t \e n_1$ and $c'$ contains at least one literal of the form $t \e n_2$.
        Let $\circ$ be the involution such that $n^\circ$ is $n^\ast$ for every $n \in \Names(c) \setminus \Names(\phi,\alpha)$ and that furthermore swaps every $n \in (\Names(\ell) \cup \Names(c')) \setminus (\Names(c) \cup \Names(\phi,\alpha))$ with $n^\circ \in \NamesPV1_n(\phi,\alpha) \setminus \Names_n(\phi,\alpha)$ and leaves the rest unchanged.
        Such an involution exists because $\Names_t(c') \supseteq (\Names_t(\ell) \setminus \{n_1\})$ and $|(\Names_t(c') \cup \{n_1\}) \setminus \Names_t(\phi,\alpha)| \leq |\NamesPV1_t(\phi,\alpha) \setminus \Names_n(\phi,\alpha)|$.
        Then $\circ$ is an involution that swaps all $n \in (\Names(c') \cup \Names(\ell)) \setminus \Names(\phi,\alpha)$ with $n^\circ \in \NamesPV1_n(\phi,\alpha) \setminus \Names_n(\phi,\alpha)$.
        By induction, $c'{}^\circ, \ell^\circ \in \UP[k]{\gnd_{\NamesPV1(\phi,\alpha)}(\phi)}$.
        Since $c^\ast$ is the unit propagation of $c'{}^\circ$ and $\ell^\circ$, we have $c^\ast \in \UP[k+1]{\gnd_{\NamesPV1(\phi,\alpha)}(\phi)}$.
        \qedhere
    \end{enumerate}
\end{itemize}
\end{pf}

\subsection{Decision Procedure}

\begin{defi}
Let $\phi$ be proper\plus\ and $\psi$ be objective.
Let $\phi|_T$ be the conjunction of clauses of the form $\bigvee_{1 \leq i \leq j, n \notin \NamesP(\phi,\psi)} x_i \e n \lor \bigvee_{i_1 \neq i_2} x_{i_1} \e x_{i_2} \lor c^{n_1 \ldots n_j}_{x_1 \ldots n_j}$ for every $c \in \gnd_{\NamesPV1(\phi,\psi)}(\phi)|_T$ where $x_1,\ldots,x_j$ are of the same sorts as $\{n_1,\ldots,n_j\} = \NamesPV1(\phi,\psi) \setminus \Names(\phi,\psi)$, and where the empty clause is turned into a clause $x \n x$.
\end{defi}

\begin{lem} \label{lem:setup-g-restrict}
Let $\phi$ be proper\plus, $\psi$ be objective.
Let $N \supseteq \NamesPV1(\phi,\psi)$.
Then $\WP{\gnd_N(\phi|_T)} = \gnd_N(\phi)|_T$.
\end{lem}

\begin{pf}
For the $\subseteq$ direction suppose $c' \in \WP{\gnd_N(\phi|_T)}$.
Then $c'$ is the result of unit propagation with the grounding of a formula $\bigvee_{1 \leq i \leq j, n \notin \NamesP(\phi,\psi)} x_i \e n \lor \bigvee_{i_1 \neq i_2} x_{i_1} \e x_{i_2} \lor c^{n_1 \ldots n_j}_{x_1 \ldots n_j}$ with names $n'_1,\ldots,n'_j$.
Then $n'_i \in N \setminus \Names(\phi,\psi)$ and $n'_{i_1} \neq n'_{i_2}$, for otherwise the clause were subsumed by the valid literal $n'_{i_1} \e n'_{i_2}$ and hence not in $\WP{\gnd_N(\phi|_T)}$.
Thus $c'$ is just $c^{n_1 \ldots n_j}_{n'_1 \ldots n'_j}$; for if $c'$ were the result unit propagation of $c^{n_1 \ldots n_j}_{n'_1 \ldots n'_j}$, then $c'$ would subsume $c^{n_1 \ldots n_j}_{n'_1 \ldots n'_j}$, and by Lemma~\ref{lem:setup-gnd-finite}, $c'{}^{n'_1 \ldots n'_j}_{n_1 \ldots n_j} \in \UP{\gnd_N(\phi)}$ and $c \notin \WP{\gnd_N(\phi)}$ and hence also $c \notin \gnd_N(\phi)|_T$, which contradicts the assumption.
Then by Lemma~\ref{lem:setup-gnd-finite} $c' \in \WP{\gnd_N(\phi)}$, and an induction on the construction of $\gnd_N(\phi)|_T$ shows that $c' \in \gnd_N(\phi)|_T$.

Conversely, suppose $c \in \gnd_N(\phi)|_T$.
Then $\phi|_T$ contains a clause $c'$ of the form $\bigvee_{1 \leq i \leq j, n \notin \NamesP(\phi,\psi)} x_i \e n \lor \bigvee_{i_1 \neq i_2} x_{i_1} \e x_{i_2} \lor c^{n_1 \ldots n_j}_{x_1 \ldots n_j}$.
Then $c'{}^{x_1 \ldots x_j}_{n_1 \ldots n_j} \in \UP{\gnd_{\NamesPV1(\phi,\psi)}(\phi|_T)}$, and by unit propagation with valid literal clauses, $c \in \UP{\gnd_{\NamesPV1(\phi,\psi)}(\phi|_T)}$.
Then $c$ is not subsumed by any other in $\UP{\gnd_{\NamesPV1(\phi,\psi)}(\phi|_T)}$, for otherwise $c \notin \gnd_N(\phi)|_T \subseteq \WP{\gnd_N(\phi)}$.
Thus $c \in \WP{\gnd_{\NamesPV1(\phi,\psi)}(\phi|_T)}$.
\end{pf}

\begin{lem} \label{lem:decidable-objective}
Let $\phi$ be proper\plus\ and $\psi$ be objective and without $\G$.
Then $\gnd(\phi)|_T \modelss \Lmod[k] \psi$ iff $\gnd(\phi|_T) \modelss \Lmod[k] \psi$ iff the following reduction is true:
\begin{enumerate}
\item \label{dec:ll:lit} $\gnd(\phi|_T) \modelss \ell$ iff
    $\ell$ is valid or subsumed by some $c \in \UP{\gnd_{\NamesPV1(\phi|_T,\ell)}(\phi|_T)}$
\item \label{dec:ll:or} $\gnd(\phi|_T) \modelss (\psi \lor \chi)$ iff
    \begin{itemize}[sub]
    \item $(\psi \lor \chi)$ is valid or subsumed by some $c \in \UP{\gnd_{\NamesPV1(\phi|_T,\psi,\chi)}(\phi|_T)}$ \; if $(\psi \lor \chi)$ is a clause;
    \item $\gnd(\phi|_T) \modelss \psi$ or $\gnd(\phi|_T) \modelss \chi$ \qquad\qquad\qquad\qquad\qquad\qquad\qquad\qquad otherwise;
    \end{itemize}
\item \label{dec:ll:and} $\gnd(\phi|_T) \modelss \neg (\psi \lor \chi)$ iff $\gnd(\phi|_T) \modelss \neg \psi$ and $\gnd(\phi|_T) \modelss \neg \chi$;
\item \label{dec:ll:ex} $\gnd(\phi|_T) \modelss \ex x \psi$ iff $\gnd(\phi|_T) \modelss \psi^x_n$ for some $n \in \NamesP1_x(\phi|_T,\psi)$;
\item \label{dec:ll:fa} $\gnd(\phi|_T) \modelss \neg \ex x \psi$ iff $\gnd(\phi|_T) \modelss \neg \psi^x_n$ for every $n \in \NamesP1_x(\phi|_T,\psi)$;
\item \label{dec:ll:negneg} $\gnd(\phi|_T) \modelss \neg \neg \psi$ iff $\gnd(\phi|_T) \modelss \psi$;
\item \label{dec:ll:k0} $\gnd(\phi|_T) \modelss \K[0] \psi$ iff
    \begin{itemize}[sub]
    \item $\gnd_{\NamesPV1(\phi|_T)}(\phi|_T)$ is obviously inconsistent, or
    \item $\gnd(\phi|_T) \modelss \psi$;
    \end{itemize}
\item \label{dec:ll:kk} $\gnd(\phi|_T) \modelss \K[k+1] \psi$ iff
    \begin{itemize}[sub,nolabel]
    \item for some $t \in \Terms(\gnd_{\NamesPV0(\phi|_T,\psi)}(\phi|_T,\psi))$ and every $n \in \NamesP1_{t\bullet}(\phi|_T,\psi,t)$,
    \item $\gnd(\phi|_T \land t \e n) \modelss \K[k] \psi$;
    \end{itemize}
\item \label{dec:ll:negk} $\gnd(\phi|_T) \modelss \neg \K[k] \psi$ iff $\gnd(\phi|_T) \not\modelss \K[k] \psi$;
\item \label{dec:ll:m0} $\gnd(\phi|_T) \modelss \M[0] \psi$ iff
    \begin{itemize}[sub]
    \item $\gnd_{\NamesPV1(\phi|_T)}(\phi|_T)$ is not potentially inconsistent, and
    \item $\gnd(\phi|_T) \modelss \psi$;
    \end{itemize}
\item \label{dec:ll:mk} $\gnd(\phi|_T) \modelss \M[k+1] \psi$ iff
    \begin{itemize}[sub,nolabel]
    \item for some $t \in \Terms(\gnd_{\NamesPV0(\phi|_T,\psi)}(\phi|_T,\psi))$ and $n \in \NamesP1_{t\bullet}(\phi|_T,\psi,t)$,
    \item $\gnd(\phi|_T \land t \e n) \modelss \M[k] \psi$ or
    \item $\gnd(\phi|_T \uplus (t \e n)) \modelss \M[k] \psi$;
    \end{itemize}
\item \label{dec:ll:negm} $\gnd(\phi|_T) \modelss \neg \M[k] \psi$ iff $\gnd(\phi|_T) \not\modelss \M[k] \psi$.
\end{enumerate}
\end{lem}

\begin{pf}
The step from $\gnd(\phi)|_T$ to $\gnd(\phi|_T)$ holds by Lemmas \ref{lem:setup-g-restrict} and \ref{lem:setup-fix-modelss}.
Rules \ref{dec:ll:lit} and \ref{dec:ll:or} follow from Lemma~\ref{lem:setup-gnd-finite}.
Rule~\ref{dec:ll:and} is trivial.
Rules \ref{dec:ll:ex} and \ref{dec:ll:fa} follow from Lemma~\ref{lem:lmod-objective-involution-kb-unchanged}.
Rule~\ref{dec:ll:negneg} is trivial.
Rule~\ref{dec:ll:k0} follows from Lemma~\ref{lem:setup-gnd-finite}.
Rule~\ref{dec:ll:kk} follows from Lemma~\ref{lem:split-objective-k}.
Rule~\ref{dec:ll:negk} is trivial.
Rule~\ref{dec:ll:m0} follows from Lemma~\ref{lem:setup-gnd-finite}.
Rule~\ref{dec:ll:mk} follows from Lemma~\ref{lem:split-objective-m}.
Rule~\ref{dec:ll:negm} is trivial.
\end{pf}

\begin{repthm}[Decidability]{thm:decidable}
Let $\phi$ be proper\plus\ and $\sigma$ be objective.
Then $\gnd(\phi) \modelss \sigma$ is decidable.
\end{repthm}

\begin{pf}
Follows directly from Theorem~\ref{thm:representation} and Lemma~\ref{lem:decidable-objective}.
\end{pf}

\subsection{Complexity}

This section proves the complexity results Theorem~\ref{thm:tractable-objective} and Corollary~\ref{cor:tractable}.

\begin{repthm}[Tractability]{thm:tractable-objective}
Let $\phi, \psi$ be propositional, $\phi$ be proper\plus, $\psi$ be objective.
Then $\OO \phi \entailss \Lmod[k] \psi$ is decidable in $\BigO(2^k (|\phi| + |\psi|)^{k+2})$.
\end{repthm}

\begin{pf}
First consider $\OO \phi \entailss \K[k] \psi$.
Let $f(k)$ be the complexity of deciding this with the procedure from Lemma~\ref{lem:decidable-objective} after $k$ splits.
Then $f(0) \in \BigO((|\phi| + k) \cdot |\psi|)$, as at most $k$ literals have been added to $\phi$ and unit propagation and the at most $|\psi|$ subsumptions can be computed in linear time.
Furthermore, $f(k+1) \in \BigO((|\phi| + |\psi|) \cdot f(k))$ as only primitive terms and their corresponding right-hand side plus one more name need to be tested for splitting; there are at most $\tfrac13 \cdot (|\phi| + |\psi|)$ primitive terms in $\phi$ and $\psi$.
Then $f(k) \in \BigO((|\phi| + |\psi|)^k \cdot (|\phi| + k) \cdot |\psi|) = \BigO((|\phi| + |\psi|)^k \cdot (|\phi| + |\phi| + |\psi|) \cdot (|\phi| + |\psi|)) = \BigO((|\phi| + |\psi|)^k \cdot 2 \cdot (|\phi| + |\psi|) \cdot (|\phi| + |\psi|)) = \BigO((|\phi| + |\psi|)^{k+2})$.

Now consider $\OO \phi \entailss \M[k] \psi$.
Let $f(k)$ be the complexity of deciding this with the procedure from Lemma~\ref{lem:decidable-objective} after $k$ splits.
Then $f(0) \in \BigO(2 \cdot (|\phi| + |\psi|) \cdot |\psi|)$, as at most $|\phi| + |\psi|$ literals have been added during splitting.
Furthermore, $f(k+1) \in \BigO(2 \cdot (|\phi| + |\psi|) \cdot f(k))$ as only terms in $\phi$ and $\psi$ are split with their corresponding right-hand side plus one more name, and both the one-literal and the isomorphic-literals variants need to be checked; there are at most $\tfrac13 \cdot (|\phi| + |\psi|)$ primitive terms in $\phi$ and $\psi$.
Then $f(k) \in \BigO(2^k \cdot (|\phi| + |\psi|)^k \cdot 2 \cdot (|\phi| + |\psi|) \cdot |\psi|) = \BigO(2^k \cdot (|\phi| + |\psi|)^{k+1} \cdot |\psi|) = \BigO(2^k \cdot (|\phi| + |\psi|)^{k+2})$.
\end{pf}

\begin{repcor}[Tractability]{cor:tractable}
Let $\phi, \sigma$ be propositional, $\phi$ be proper\plus, $\sigma$ be subjective, and $k \geq l$ for every $\K[l], \M[l]$ in $\sigma$.
Then $\OO \phi \entailss \sigma$ is decidable in $\BigO(2^k (|\phi| + |\sigma|)^{k+3})$.
\end{repcor}

\begin{pf}
$\sigma$ contains at most $|\sigma|$ subformulas $\Lmod[l] \sigma'$, each of which by Theorem~\ref{thm:representation} reduces to at most $|\sigma|$ instances of solving $\OO \phi \entailss \Lmod[l] \psi$ where $|\psi| \leq |\sigma'| \leq |\sigma|$.
By Theorem~\ref{thm:tractable-objective}, each instance can be solved in $\BigO(2^k (|\phi| + |\sigma|)^{k+2})$.
\end{pf}

\end{appendices}

\end{document}